%% file: main.tex
\documentclass[sigconf]{acmart}
\usepackage{algorithm} 
\usepackage{algorithmicx} 
\usepackage{algpseudocode} 
\usepackage{amsmath} 
\usepackage{threeparttable}
\usepackage{graphicx}
\usepackage{hyperref}
\usepackage{subcaption}
\AtBeginDocument{%
 \providecommand\BibTeX{{%
  \normalfont B\kern-0.5em{\scshape i\kern-0.25em b}\kern-0.8em\TeX}}}

\begin{document}

\title{CD-SGD: Distributed Stochastic Gradient Descent with Compression and Delay Compensation}
\author{Enda Yu, Dezun Dong, Yemao Xu, Shuo Ouyang, Xiangke Liao}
\authornote{Dezun Dong is the corresponding author.}
\affiliation{%
  \institution{College of Computer, National University of Defense Technology}
  \city{Changsha 410073}
  \country{China}
  }
\email{{yuenda,dong,xuyemaovip,ouyangshuo,xkliao}@nudt.edu.cn}

\begin{abstract}
Communication overhead is the key challenge for distributed training. Gradient compression is a widely used approach to reduce communication traffic. When combining with parallel communication mechanism method like pipeline, gradient compression technique can greatly alleviate the impact of communication overhead. However, there exists two problems of gradient compression technique to be solved. Firstly, gradient compression brings in extra computation cost, which will delay the next training iteration. Secondly, gradient compression usually leads to the decrease of convergence accuracy. In this paper, we combine parallel mechanism with gradient quantization and delayed full-gradient compensation, and propose a new distributed optimization method named CD-SGD, which can hide the overhead of gradient compression, overlap part of the communication and obtain high convergence accuracy. The local update operation in CD-SGD allows the next iteration to be launched quickly without waiting for the completion of gradient compression and current communication process. Besides, the accuracy loss caused by gradient compression is solved by k-step correction method introduced in CD-SGD. We prove that CD-SGD has convergence guarantee and it achieves at least 
\begin{math}
 O(\frac {1}{\sqrt{K}}+\frac {1}{K})
\end{math} convergence rate. We conduct extensive experiments on MXNet to verify the convergence properties and scaling performance of CD-SGD. Experimental results on a 16-GPU cluster show that convergence accuracy of CD-SGD is close to or even slightly better than that of S-SGD, and its end-to-end time is 30$\%$ less than 2 bit gradient compression under 56Gbps bandwidth environment.
\end{abstract}



\maketitle

\section{Introduction}

Distributed training has become an effective method for deep learning model training. The enormous training data set is divided among multiple nodes for training tasks. As a result, these nodes must communicate their calculated parameters with each other before updating global parameters. The communication cost limits the scalability of the distributed system and reduces the efficiency of distributed training seriously. For example, when training ResNet-50 on a 16-node Nvidia P102-100 GPU cluster connected by 1Gbps Ethernet \cite{shi2020communication}, communication time is more than nine times the computation time. The communication cost tends to worsen when the number of workers increases. To address the communication challenge, many approaches are proposed to speed up distributed training, which can be divided into system-level approaches and algorithm-level approaches.

On the system level, pipelining \cite{2017S, li2018pipe, shi2019mg, zhang2017poseidon} is firstly introduced based on the layer-wise structure of neural network, which enables every back-propagation (BP) to overlap the communications with computation process of the next layer. After pipelining, communication priority scheduling mechanism \cite{hashemi2018tictac, peng2019generic} is proposed to achieve a more aggressive overlap ratio between computation and communication overhead. Recently work \cite{wang2019scalable,lin2018don} improve the distributed training performance by parallelizing the computation and communication operations. Post-local SGD \cite{lin2018don} , K-AVG \cite{2018On} and Periodic Averaging \cite{haddadpour2019local} makes every worker evolve a local model by performing local updates before communication (synchronization by averaging).

On the algorithm level, gradients compression techniques are proposed to cut down communication traffic, which can be divided into gradient sparsification \cite{strom2015scalable, aji2017sparse,lin2017deep,chen2018adacomp,fang2019redsync} and gradient quantization \cite{seide20141, alistarh2017qsgd, wen2017terngrad, bernstein2018signsgd,karimireddy2019error,ouyang2021communication}. Gradient quantization transforms high-precision gradients into low-precision ones to communicate. 1-bit quantization \cite{seide20141} reduces the communication traffic by encoding the 32-bit gradients to 1 bit. QSGD \cite{alistarh2017qsgd} allows users to choose different degrees of quantization according to network bandwidth. WAGE \cite{wu2018training} and 8-bit training \cite{banner2018scalable} quantizes not only gradients but also weights. Earlier sparsification methods \cite{strom2015scalable, aji2017sparse} judge whether to send gradients by a single threshold. Then, DGC \cite{lin2017deep} further accelerates large-scale distributed training by only exchanging top 0.1\% gradients in each iteration and accumulating the other gradients until they become large enough. Although these communication algorithms can relieve the pressure of communication, they introduce extra computation overheads on data coding and selection of gradients. Even worse, when extra computation overhead and gradient computation time are much higher than communication cost, the performance improvement of compression methods is not obvious.

Is there an appropriate way to combine the advantages of system-level approaches and algorithm-level approaches? A few studies have made efforts in this area, but the effect is still not satisfactory. LAGS-SGD \cite{shi2019layer} integrates DGC with pipelining, while it does not bring great speed advantage as there exists the startup cost of many-layer communications and extra compression cost. Canary \cite{zhou2020canary} combines 8-bit quantization with gradient partition, while it cannot solve the problem of precision decline. OMGS-SGD \cite{shi2020communication} combines DGC with the optimal merged mechanism, while it is not significantly faster than DGC when training communication intensive models such as VGG-16. Generally, there are three challenges need to be handled. Firstly, an appropriate method is needed to eliminate or cover up the extra cost of compression. Secondly, the decrease of accuracy caused by compression needs to be solved. Finally, the selected mechanism optimization method must bring enough training efficiency benefits. 

To tackle the challenges mentioned, we propose distributed stochastic gradient descent with compression and delay compensation (CD-SGD). CD-SGD explore gradient quantization and delayed full-gradient compensation to accelerate the distributed training while preserving convergence accuracy. 
% We also come up with a model to analyze when CD-SGD can achieve obvious 
% advantages than simple compression or parallel method, and the modeling results show that CD-SGD performs better when training with communication intensive model. 
We evaluate our proposed algorithm on different DNNs and verify its convergence property. Experimental results show that CD-SGD can speed up to 30$\%$ than 2bit quantization and its convergence accuracy is close to or even better than that of synchronous stochastic gradient descent (S-SGD). The contributions of our work are summarized as follows:
\begin{itemize}
\item We combine 2 bit quantization in MXNet with parallel mechanism to cover the extra quantization cost. On this basis, we propose a new distributed optimization algorithm named CD-SGD. The proposed algorithm enables us to embrace the benefits of both communication overhead concealment and gradient compression.
\item We design a periodic accuracy correction method named k-step correction to solve the problem of accuracy degradation caused by gradient compression.
\item We design a mathematical model to analyze when CD-SGD can achieve an obvious performance advantage, and we provide a mathematical proof of the convergence of CD-SGD.
\item We conduct extensive experiments on a 16-GPU cluster to verify the convergence of CD-SGD, and we compare the performance of CD-SGD with that of various algorithms.
\end{itemize}

The rest of the paper is organized as follows. Section \ref{II} introduces the preliminaries and the motivation of our work. We provide the detailed design and implement of CD-SGD algorithm in Section \ref{4}. This section also provides proof of the convergence of CD-SGD. Section \ref{5} demonstrates the speed superiority of CD-SGD and its convergence accuracy through experimental results. Section \ref{6} introduces related work, and the conclusion of this paper is in Section \ref{7}.

\begin{figure}[h]
 \centering
 \includegraphics[width=\linewidth]{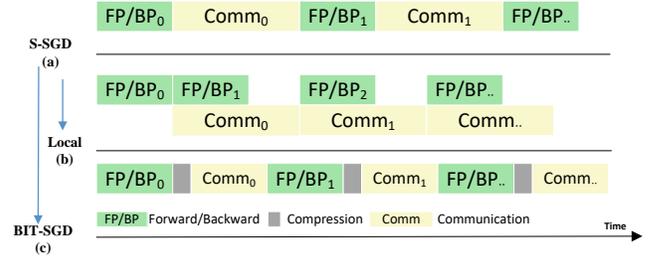}
 \caption{ Flow chart of evolution from S-SGD to OD-SGD and BIT-SGD. 
}
\label{fig1}
\end{figure}
\section{PRELIMINARY AND MOTIVATION}
\label{II}
In this section, we introduce the process of S-SGD, local update mechanism (a parallel mechanism) and gradient quantization. Besides, we point out the advantages and disadvantages of local update mechanism and gradient quantization to help understand why we want to combine them. For ease of presentation, Table \ref{tab:freq} lists the frequently used notations throughout this paper.

\begin{table}[h]
 \caption{The involved variables and their definitions}
 \label{tab:freq}
 \begin{tabular}{ll}
  \toprule
  {\bf \small Symbol}&\qquad {\bf\small Meaning}\\ 
  \midrule
  $FP_i$ & \begin{tabular}[c]{@{}l@{}}Forward propagation in $i^{th}$ iteration\end{tabular}\\ 
  $BP_i$ & \begin{tabular}[c]{@{}l@{}}Backward propagation in $i^{th}$ iteration\end{tabular}\\   
  $W_i^{loc}$   & \begin{tabular}[c]{@{}l@{}}The local weight used in FP/BP in $i^{th}$ iteration\end{tabular}\\
  $W_{i+1}$     & \begin{tabular}[c]{@{}l@{}}The global weight produced in $i^{th}$ iteration\end{tabular}\\
  $\tau$   & \begin{tabular}[c]{@{}l@{}}The computation time per iteration\end{tabular}\\
  $\varphi$   & \begin{tabular}[c]{@{}l@{}}The uncompressed communication time per iteration\end{tabular}\\
  $\psi$   & \begin{tabular}[c]{@{}l@{}}The compressed communication time per iteration\end{tabular}\\  
  $\delta$   & \begin{tabular}[c]{@{}l@{}}The extra time brought by compression\end{tabular}\\

 \bottomrule
\end{tabular}
\end{table}

\subsection{Synchronous Stochastic Gradient Descent}\label{21}
S-SGD is widely used in distributed deep learning because of good convergence properties. However, S-SGD requires the faster worker nodes to wait for the slower ones to communicate their information per iteration. This mechanism often leads to enormous communication, which greatly restricts the speed of training. As shown in Fig.\ref{fig1}a, S-SGD contains several main steps in every iteration. After loading a batch-size of data, the worker node starts forward propagation ($FP$) to calculate the loss value. Then the worker node calculates the gradient in the process of backward propagation ($BP$). The gradient will be sent to the server node. After the server node gets the gradients from all worker nodes, it begins to update global weights. In equation \ref{eq1} (eq.\ref{eq1}), the relationship between above parameters is shown, in which ${\triangledown}L(W_i;D_j)$ is the gradient from $j^{th}$ worker node and $\eta$ is an hyperparameter called learning rate used to adjust the influence of gradients on weights. 
\begin{figure*}[h]
 \centering
 \includegraphics[width=\linewidth]{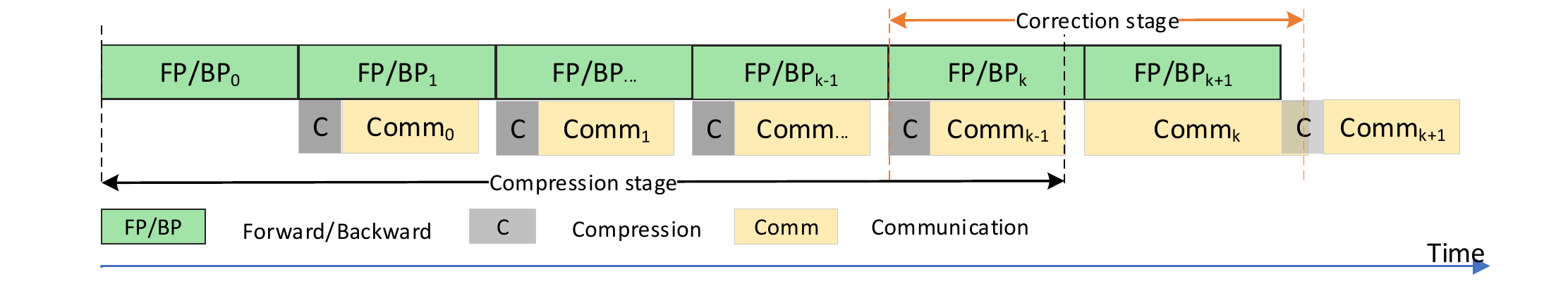}
 \caption{ Flow chart of CD-SGD. Numbers in this figure
 represents the iteration, and it shows the case when $k$ is equal to 4 .
}
\label{cd}
\end{figure*}

\begin{equation}
 W_{i+1} = W_i - {\frac{\eta}{N}{\sum_{j=1}^N}{\triangledown}L(W_{i};D_j)}
 \label{eq1}
\end{equation}

When the update operation is finished, the worker nodes pull the updated weight $W_{i+1}$ from server nodes and then start the next $FP$. S-SGD constantly repeats this process to train a mature model for application. As batch size is constant and the model load data randomly, the total time cost of $FP$ and $BP$ in every iteration can be seen as stable, which is represented by $\tau$. The communication cost is affected by the size of the parameters. It is also stable, which is represented by $\varphi$. Then the average cost of S-SGD in one iteration can be assessed as eq.\ref{eq2}.

\begin{equation}
  T^{ssgd} = \tau + \varphi   
  \label{eq2}
\end{equation}

\subsection{Local Update Mechanism}

At present, the local update mechanism is already a mature parallel mechanism, and some work \cite{lin2018don,2018On,haddadpour2019local,xu2020od} is based on it for distributed communication optimization. Local update mechanism evolves a local model on each worker to better balance the available system resources. In every iteration, the local device copies the global weight into the local model as local weight after communication (eq.\ref{eq3}).
\begin{equation}
\label{eq3}
  W_{i+1}^{loc} = W_i    
\end{equation}

The local weight is updated by the local gradients produced in next iteration, and then it will replace the global weight to participate in forward propagation. The update of local weight starts with communication operation in parallel, so it will not delay the training process. Fig.\ref{fig1}b shows the main process of local update mechanism, in which non-time-consuming operations are not shown. It can be noticed that if communication in $i^{th}$ iteration is not finished, the worker nodes cannot start $FP$ in ${i+2}^{th}$ iteration because the local weight is not ready. That means the wait time between $BP_i$ and $FP_{i+1}$ cannot be avoided if the communication cost is greater than the computation cost. Therefore, the iteration time of local update mechanism can be calculated by $\tau$ and $\varphi$ in eq.\ref{eq4}.
\begin{equation}
T^{loc} =
\begin{cases} 
\tau    & \tau > \varphi \\
\varphi    & \tau < \varphi
\end{cases}
\label{eq4}
\end{equation}

\subsection{Gradient Quantization}

As mentioned in \ref{21}, gradient aggregation communication between nodes brings huge communication overhead, which seriously affects the efficiency of training large neural networks. Generally, we can use gradient quantization to reduce the communication traffic to improve training efficiency. Currently, most machine learning platforms provide the implementation of gradient quantization. The idea of gradient quantization is to replace the original floating-point precision with lower precision (such as 8-bit integer). In this paper, various gradient quantization methods are represented by BIT-SGD. The process of BIT-SGD is shown in Fig.\ref{fig1}c. After gradients are produced, the workers encode the 32-bit gradients into lower bit data. The process of quantization delays the communication but the average communication cost can be shortened. Like S-SGD, BIT-SGD cannot start the next iteration until the communication is finished, which restrict its training efficiency. In other words, If the total time of the extra quantization cost and the optimized communication is greater than the original communication time, the quantification will bring negative benefits instead. The training cost of BIT-SGD in one iteration is shown by eq.\ref{eq5}, in which the extra compression time is indicated by the symbol $\delta$, and the optimized communication cost can be denoted by $\psi$. 
\begin{equation}
  T^{bit}= \tau +\delta + \psi 
  \label{eq5}
\end{equation}

In BIT-SGD, the errors between original gradients and the compressed ones are saved in a residual buffer. Then the data in residual buffer will be sent to server nodes when it reaches the threshold. Therefore, the value of threshold has a great influence on the accuracy. The values of parameters in various models are different, and they are usually not uniform enough. That means although 2-bit quantization preserves small gradients to ensure no information loss, it can lead to some undesirable situations. For example, only when the retained residuals accumulate beyond the threshold can they participate in the weight update, which causes some weights remaining little changed for a long time. Sometimes modifying the threshold may help to solve this problem, but various models have different parameter characteristics, and it is difficult to find a suitable threshold for them. 

According to the previous analysis, we could get a conclusion that it is appropriate to combine local update mechanism and gradient quantization. In this way, the problem that local update mechanism has a poor acceleration effect when the communication cost is significantly greater than the computation cost is solved, and gradient quantization can have better parallelism. Besides, a method to deal with the problem of accuracy degradation caused by quantization is in need. We want to solve these problems and provide a solution.

\section{DESIGN AND IMPLEMENTATION}
\label{4}
In this section, we provide a design overview about CD-SGD. After that, we introduce the implementation of CD-SGD on MXNet and give a performance comparison chart of CD-SGD and BIT-SGD recorded by profiler in MXNet. Besides, we show the effect of our method on performance improvement and provide its theoretical performance limits through time cost modeling analysis. Finally, we prove that our algorithm achieves at least 
\begin{math}
 O(\frac {1}{\sqrt{K}}+\frac {1}{K})
\end{math} convergence rate.

% we use the execution process of each operation during training measured by the profiler in MXNet to illustrate the effect of CD-SGD in quantization overhead hiding. 
\subsection{Design Overview}
\label{over}
\begin{figure}[t]
 \centering
 \includegraphics[width=\linewidth]{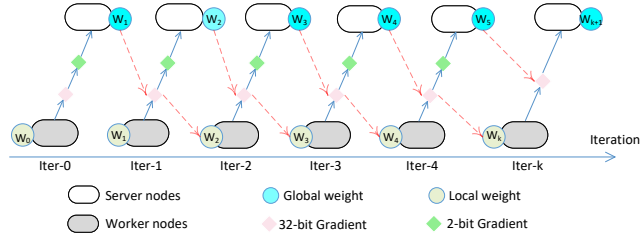}
 \caption{ The design diagram of CD-SGD. The red arrow is used to describe the local weight generation process, and the blue arrow is used to describe the gradient transfer process.}
\label{ks}
\end{figure}

In this paper, we focus on the 2-bit quantization in MXNet, and according to its mechanism characteristics, we introduce the local update into MXNet to combine with it. Besides, we design a method named k-step correction to reduce the impact of accuracy degradation caused by the delayed updates. The working process of CD-SGD is shown in Fig.\ref{cd}, which consists of compression stage and correction stage, being used to accelerate the training and improve training accuracy respectively. In the iteration of compression stage, worker nodes calculate gradients in process of FP/BP, and then the gradients are used to update the local weight and do the quantization. After the local update is finished, the computation of next iteration could start. And it could work with current communication in parallel. In correction stage, worker nodes perform no quantization operation, which is different from the compression stage. The specific design is introduced as follows.

As mentioned in Section \ref{II}, the quantization operation delays the update of global weights. And the global weights are used in next iteration to calculate new parameters. In order to start the next round of training faster, the local update operation is introduced to change the dependence of the next iteration on the global weights. As shown in Fig.\ref{ks}, CD-SGD sets a local buffer, which loads global weights in last iteration as local weights. The local weights are updated by the local gradients produced in current iteration and then replace the global weights to participate computation in next iteration. So that the computation in next iteration could start without waiting the end of current communication. 

The design logic of k-step correction is using 32-bit gradient communication periodically to help global weights to correct in time in the right direction. The value of $k$ can adjust the number of iterations of the compressed state in CD-SGD. In every $k$ iterations, there are $k-1$ iteration in which worker nodes encode the 32-bit gradients into 2-bit gradients to reduce communication traffic. The rest of the iteration is the correction state. As shown in Fig.\ref{ks}, no matter it is the compression state or the correction state, the local weights are updated with 32-bit gradients, which makes the calculation more stable and further reduces the impact of the compressed gradient on the accuracy. Although not all weight updates need to work immediately, the improvement of accuracy cannot perform very well when $k$ is too big. Moreover, the average iteration time of CD-SGD will be influenced by the value of $k$. However, the influence of $k$ on speed also depends on the environment. Therefore, the suitable $k$ is much more like an empirical trick and we provide the experimental results of different $k$ when training the same model in Section \ref{5}.

\begin{figure}[t]
 \centering
 \includegraphics[width=\linewidth]{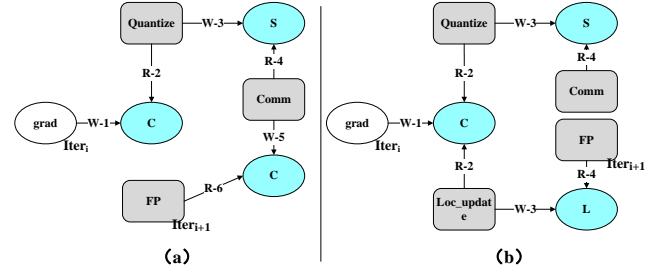}
 \caption{Implementation of BIT-SGD in (a) and CD-SGD in (b). Numbers at the end of R and W describe the order of read and write. \texttt{$C$}, \texttt{$L$}, \texttt{$S$} are \texttt{$comm\_buf$}, \texttt{$sml\_buf$} and \texttt{$loc\_buf$} respectively. }
\label{bx}
\end{figure}

\subsection{Implementation}
In this subsection, we introduce the implementation of CD-SGD based on the PS architecture on MXNet. And we compare the implementation of BIT-SGD and CD-SGD to help readers further understand our work. Besides, we show the performance comparison chart of BIT-SGD and CD-SGD record by profiler of MXNet to verify the implementation effect.

\subsubsection{Implementation on MXNet}
We implement CD-SGD based on PS architecture in MXNet. And we provide the process of CD-SGD working on the worker node in Algorithm \ref{suanfa}, which contains warm up phase and formal training phase. The warm up phase is used to stabilize weights quickly, and the formal training phase is the implementation of CD-SGD diagram in Fig.\ref{ks}. 

In warm up phase, there are $n$ iterations. In every iteration, the workers calculate the local gradients with global weights. The gradients are placed in \texttt{$comm\_buf$} and then be pushed to the server nodes to update the global weights. The updated weights will overlay previous content of \texttt{$comm\_buf$}, so only one buffer is enough for this work. In the ${n-1}^{th}$ iteration, the global weights are copied into \texttt{$loc\_buf$}. Data in the \texttt{$loc\_buf$} is updated with local gradients in the ${n}^{th}$ iteration, which provides the necessary weights to start formal training phase. The length of warm up phase can be adjusted according to the complexity of different models, and it takes little time usually. 

In formal training phase, the worker node will copy the global gradient into the \texttt{$loc\_buf$} after pulling the global weight. The data in \texttt{$loc\_buf$} is updated with local gradients and then participates gradients calculation in next iteration. The dependency engine of MXNet can ensure that the old local weight is overwritten after it participates in the computation. Besides, a new counter is used to determine whether to perform quantization. The update of local weights and quantization operation can work in parallel because they read \texttt{$comm\_buf$} without modifying it. After gradients are encode into 2-bit data, they will be pushed to server nodes. The server nodes must decode the quantified gradients into 32 bits before updating global weights. If quantization is not needed in current iteration, the process is same as that in warm up phase. As introduced in \ref{over}, when the update of the local weight is finished, the next round can be executed immediately. This mechanism ensures that computation in next iteration cannot be delayed by compression and work with communication in parallel.\footnote{We will put the address of our source code on \href{https://github.com/Tugraph/CD-SGD}{https://github.com/Tugraph/CD-SGD}}

%\floatname{algorithm}{Algorithm} 
\renewcommand{\algorithmicrequire}{\textbf{Input:}} 
\renewcommand{\algorithmicensure}{\textbf{Initialize:}} 
 \begin{algorithm}[h]
  \caption{CD-SGD at worker $g$} 
  \begin{algorithmic}[1] %每行显示行号 
   \Require Warm-up steps $n$, k-step $k$ 
      \Ensure Counter $i$, $count$
      \Function {WARMUP}{} 
        % \State i←0 
        \While{$i<=n$} 
          \State calculate $loss_i$ with $W_{i}$
          \State calculate $grad_i^{loc}$ with $loss_i$
          \If{$i$ == $n$} 
          \State Update $W_{n+1}^{loc}$ with $grad_i^{loc}$ 
          \EndIf 
          \State Push $grad_i^{loc}$ to Server
          \State Update $W_{i}$ with global $grad_i$ on Server
          \State Pull $W_{i+1}$ from Server
          \If{$i == n-1$} 
            \State $W_{n+1}^{loc}$ ← $W_{n}$
          \EndIf     
          \State $i$←$i+1$
        \EndWhile 
      \EndFunction 
      \State 
      \Function{FormalTraining}{} 
        % \State count←0 
        \While{not stop} 
          \State calculate $loss_i$ with $W_{i}^{loc}$ 
          \State calculate $grad_i^{loc}$ with $loss_i$
          \State Update $W_{i+1}^{loc}$ with $grad_i^{loc}$ 
        \If{$count$ \% $k$ != 0} 
          \State Compress $grad_i^{loc}$ to $grad_i^s$
          \State Push compressed $grad_i^s$ to Server      
          \State Update $W_{i}$ with global $grad_i^s$ on Server    
        \Else 
          \State Push $grad_i^{loc}$ to Server
          \State Update $W_{i}$ with global $grad_i$ on Server
        \EndIf
        \State Pull $W_{i+1}$ from Server
        \State $W_{i+2}^{loc}$ ← $W_{i+1}$
        \State $i$ ← $i+1$
        \State $count$ ← $count+1$
        \EndWhile 
      \EndFunction 
    \end{algorithmic} 
    \label{suanfa}
  \end{algorithm}

\subsubsection{Quantization Overhead Hiding}
To further explain the logic of CD-SGD, we compare the implementation of BIT-SGD and CD-SGD. As for BIT-SGD in Fig.\ref{bx}a, gradients calculated is written into \texttt{$comm\_buf$}. Then worker nodes encode 32-bit gradients into 2 bit and then write them into \texttt{$sml\_buf$}. After that, the compressed gradients are sent to server nodes to update global weights. The updated global weights will be written into \texttt{$comm\_buf$} and then participate in the next FP. As for CD-SGD in Fig.\ref{bx}b, the local weights in \texttt{$loc\_buf$} are always updated with gradients in \texttt{$comm\_buf$} in the process of local update. The $FP$ in next iteration only needs to read data in \texttt{$loc\_buf$} to work properly. Obviously, the next $FP$ can work with communication and even quantization in parallel in CD-SGD. In other words, quantization overhead delays communication and has no effect on training in next iteration. For the whole training, it is hidden.
\begin{figure}[t]
  \centering
  \begin{subfigure}[b]{0.5\linewidth}
    \includegraphics[width=\linewidth]{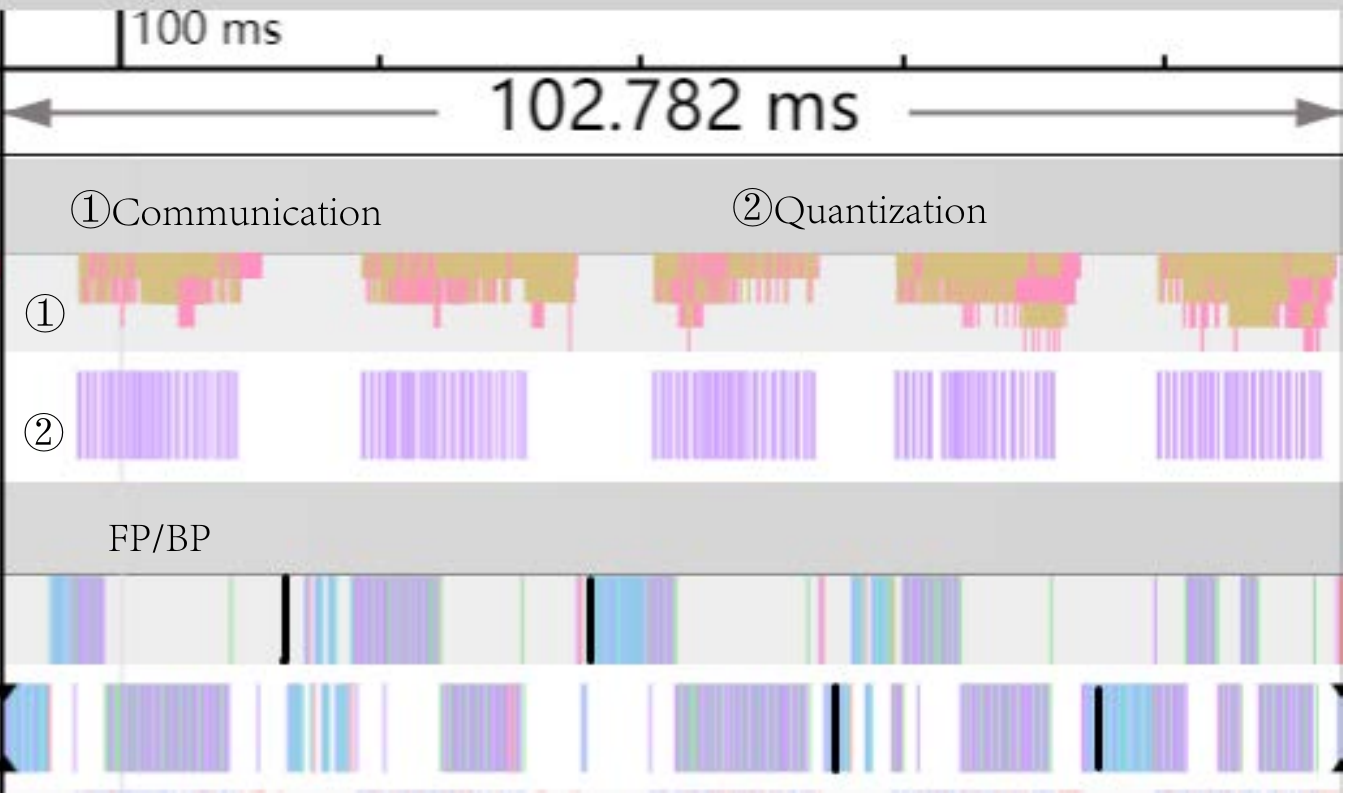}
        \caption{Tracing of BIT-SGD}
        \label{ba}
  \end{subfigure}
  ~
  \begin{subfigure}[b]{0.5\linewidth}
    \includegraphics[width=\linewidth]{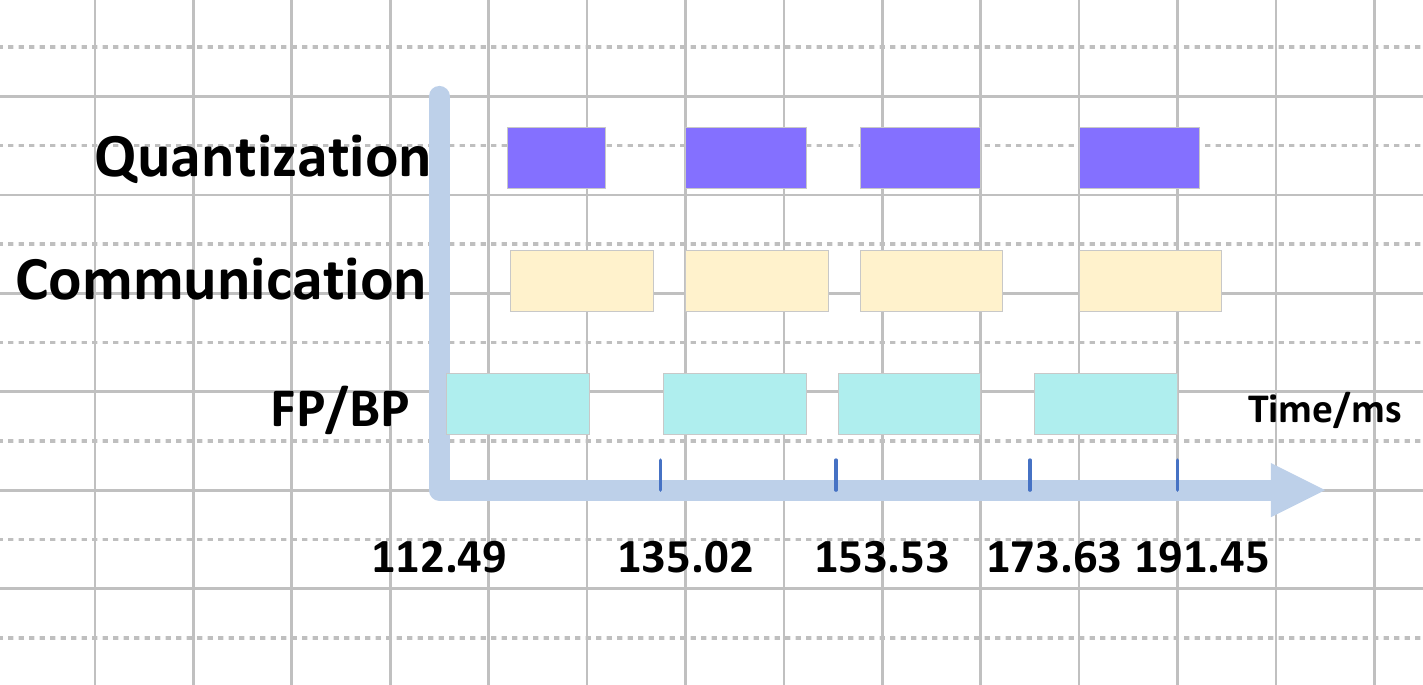}
    \caption{Timeline samples of tracing of BIT-SGD}
    \label{bb}
  \end{subfigure}
  ~
  
  \begin{subfigure}[b]{0.5\linewidth}
    \includegraphics[width=\linewidth]{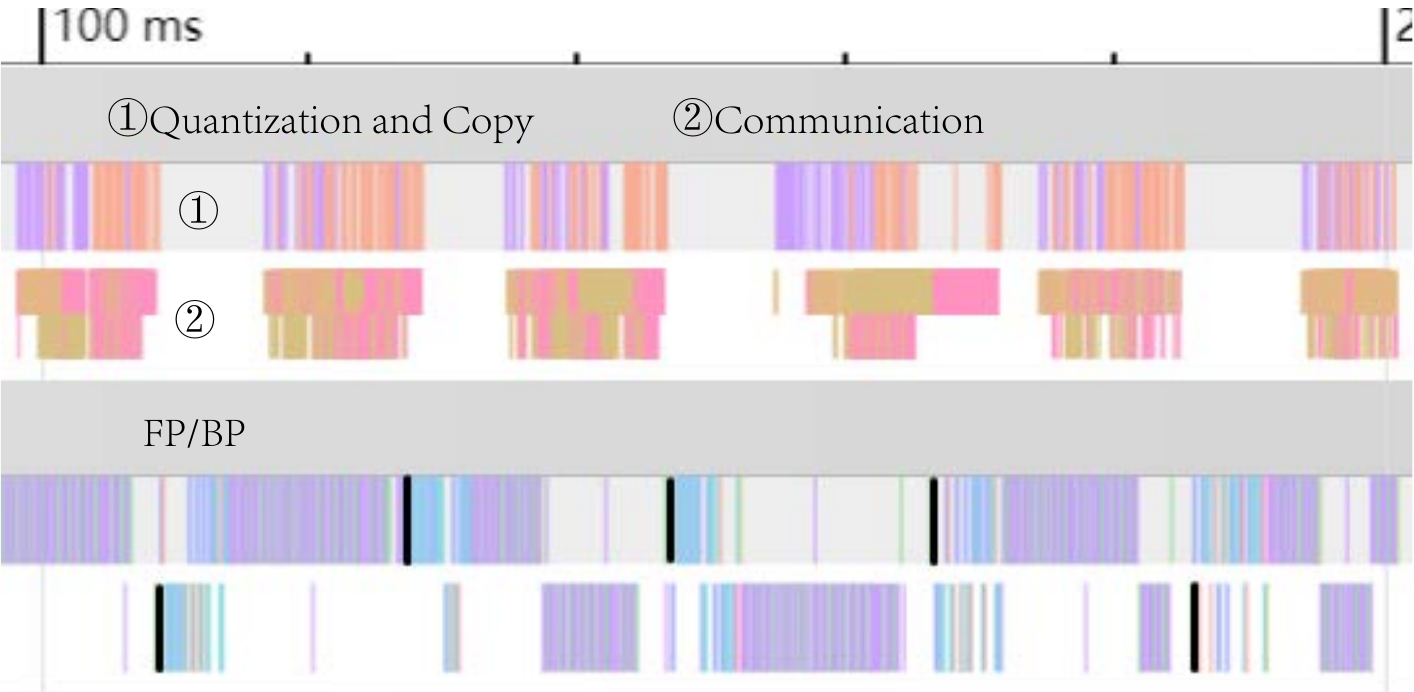}
        \caption{Tracing of CD-SGD}
  \end{subfigure}
  ~
  \begin{subfigure}[b]{0.5\linewidth}
    \includegraphics[width=\linewidth]{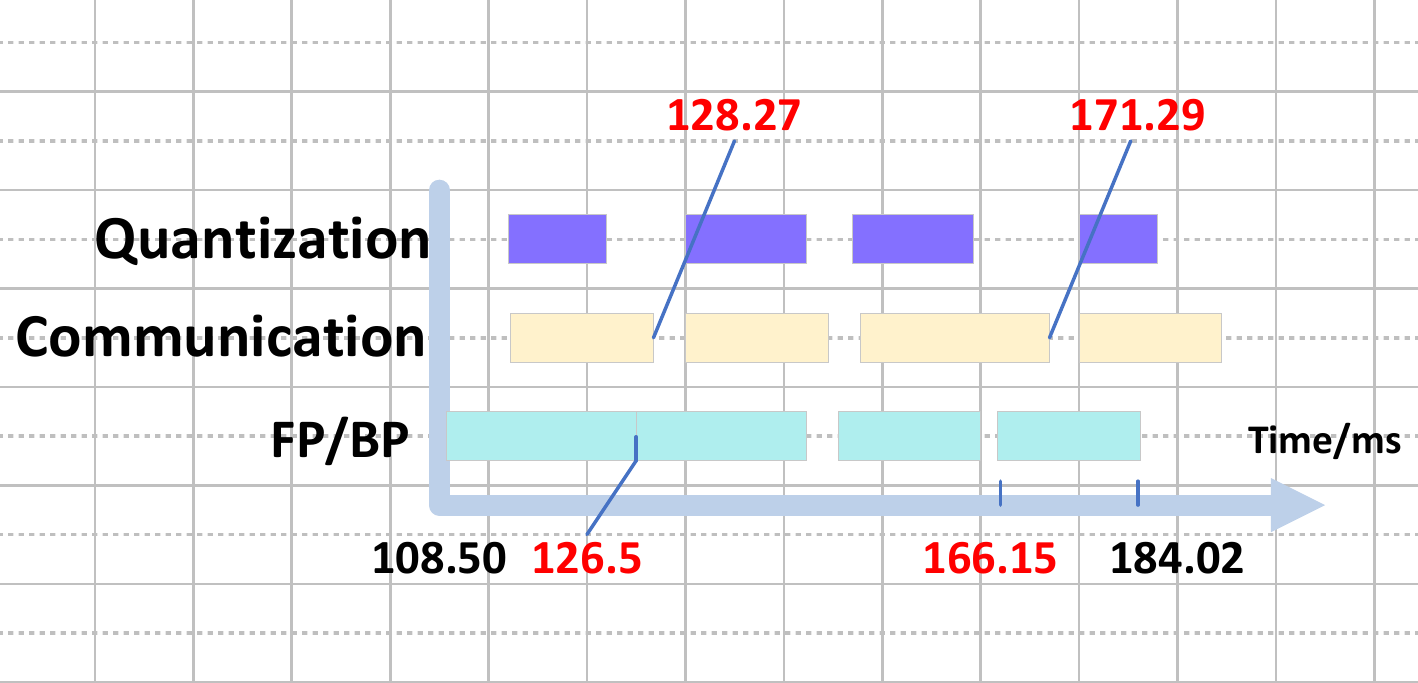}
        \caption{Timeline samples of tracing of CD-SGD}
        \label{cdb}        
  \end{subfigure}
  \caption{The execution sequence and time cost of each operation in the process of ResNet-20 model training with two worker nodes on the K80 cluster}
\label{hiding}  
\end{figure}
In order to verify whether our implementation is effective, we use the profiler in MXNet to record the execution sequence and time cost of each operation in a certain period of time during the ResNet-20 model training process. After that, we use the trace-viewer in chrome to display the obtained data. The image representation of the data obtained is shown in Fig.\ref{hiding}, which records the working data of BIT-SGD and CD-SGD between 100 and 200 milliseconds. Fig.\ref{ba} shows the execution of quantization, communication and FP/BP operations of BIT-SGD. In FP/BP area, the black vertical line is a sign of the start of $FP$ in each iteration. It helps observe whether the start of $FP$ is delayed by communication. To help understand, we provide Fig.\ref{bb} drawn based on the data in Fig.\ref{ba}. For BIT-SGD, FP/BP has to wait for the end of the communication in previous iteration to start. However, this phenomenon does not appear in Fig.\ref{cdb}. For example, the $4^{th}$ FP/BP starts at 166.15 ms, but the $3^{th}$ communication ends at 171.29 ms. As mentioned in \ref{over}, quantization delays communication. Therefore, the start time of next iteration has nothing to do with current communication, the quantization overhead can be seen as being hidden. 

Perhaps the reader will have the idea that the quantization overhead is covered by the communication cost in Fig.\ref{hiding}. In fact, the quantization and communication are carried out on a layer-by-layer basis, and each layer of quantization cannot start until its corresponding $BP$ calculates the gradient. In other words, sometimes the ${i+1}^{th}$ layer communication is completed but the ${i}^{th}$ layer quantization has not yet started. In this case, quantization cannot be parallel to communication. Therefore, the quantified cost exists objectively. The difference in time cost of the two algorithms can also prove this fact. BIT-SGD completes 5 iterations of training in 102 ms, while CD-SGD completes 6 iterations.

\subsection{Time Cost Modeling Analysis}
In this subsection, we compared the time cost of BIT-SGD and local update method with that of CD-SGD through modeling analysis. And the theoretical performance limits of CD-SGD is provided.
As mentioned above, there are k-1 compression iterations and one uncompressed iteration in every $k$ steps. For uncompressed iteration, the communication time is equal to $\varphi$. For compression iteration, the compressed communication cost is equal to $\psi$. So, the communication time in $i^{th}$ iteration of CD-SGD can be calculated in eq.\ref{eq6}.

\begin{equation}
\varphi^{cd} =
\begin{cases} 
\delta + \psi      &  Case1:{\mathit i}\ mod\ k \neq 0 \\
\varphi              &   Case2:{\mathit i}\ mod\ k = 0
\end{cases}
\label{eq6}
\end{equation}

Then the $i^{th}$ iteration time of CD-SGD can be estimated by eq.\ref{eq7}:
\begin{equation}
T^{cd} =
\begin{cases} 
\tau              & Case1:\tau > \varphi^{cd} \\
\delta + \psi        & Case2:\tau < \varphi^{cd}\ and\ {\mathit i}\ mod\ k \neq 0  \\
\varphi  & Case3:\tau < \varphi^{cd}\ and\ {\mathit i}\ mod\ k = 0
\end{cases}
\label{eq7}
\end{equation}

When compared with local update method, the saving iteration time of CD-SGD per iteration can be denoted with
\begin{math}
 T_{s}^{loc} = T^{loc} - T^{cd}.
\end{math}

\begin{equation}
T_{s}^{loc} =
\begin{cases} 
0                & Case1: \tau > \varphi   \\
\varphi - \tau       & Case2: \tau < \varphi \ and\ \tau > \varphi^{cd}  \\
\varphi - \delta-\psi      & Case3: \tau < \varphi^{cd}\ and\ {\mathit i}\ mod\ k \neq 0 \\
0      & Case4: \tau < \varphi^{cd}\ and\ {\mathit i}\ mod\ k = 0
\end{cases}
\label{eq9}
\end{equation}

When compared with BIT-SGD, the saving iteration time of CD-SGD in $i^{th}$ iteration can be denoted with
\begin{math}
 T_{s}^{bit} = T^{bit} - T^{cd}.
\end{math}

\begin{equation}
T_{s}^{bit} =
\begin{cases} 
\delta + \psi    & Case1: \tau > \varphi^{cd}  \\
\tau       & Case2: \tau < \varphi^{cd}\ and\ {\mathit i}\ mod\ k \neq 0  \\
\tau +\delta + \psi - \varphi   & Case3: \tau < \varphi^{cd}\ and\ {\mathit i}\ mod\ k = 0 
\end{cases}
\label{eq8}
\end{equation}

For local update method in eq.\ref{eq9}, the compressed communication time $\varphi^{cd}$ is less than computation time $\tau$ under normal conditions, so the saving iteration time of CD-SGD usually has two values which are 0 in case 1 and $\varphi - \tau$ in case 2. However, when the communication overhead of CD-SGD is larger than computation cost, the result will be $\varphi - \delta-\psi$ in case 3 and 0 in case 4. So, if the computation is not the bottleneck, CD-SGD is always faster than local update method. For BIT-SGD in eq. \ref{eq8}, the saving iteration time of CD-SGD is always positive in compression stage. But when multiple nodes are used to train the model which has many parameters, the saving time may be negative in case 3 of eq.\ref{eq8}. In this situation the correction stage in Fig.\ref{cd} costs more time than the average iteration time of BIT-SGD. According to Communication Model 
\cite{shi2020communication,xu2019sketchdlc}, value of the saving iteration time of CD-SGD is mainly affected by batch-size, nodes number and parameters in various training models. Therefore, we can evaluate CD-SGD according to eq.\ref{eq9} and eq.\ref{eq8}:

\textcircled{1} In the compression stage, CD-SGD can cover up the extra compression overhead of BIT-SGD and even bring better speed advantage. When training a large model and the communication cost among multiple nodes increases, an appropriate value of $k$ to maintain more iterations in compression stage is necessary for performance improvement. \textcircled{2} CD-SGD can solve the problem of local update method that it is too difficult to do communication overhead hiding when the communication time is much longer than the computation time. In this case, the theoretical average iteration time of CD-SGD is 
\begin{math}
\frac {(k-1)*(\delta + \psi)+\varphi}{k}
\end{math}.
However, when computation cost is the bottleneck of training, the acceleration effect of CD-SGD is not obvious, and in this case the difference between the training speed of CD-SGD and that of BIT-SGD is the extra compression cost.

\subsection{Proof of Convergence}
This subsection provides simple mathematical analysis of CD-SGD as well as its convergence property. Firstly, we introduce the update rules of CD-SGD to help derive the convergence formula. After that, we give the relationship between the global weight and the uncompressed gradient and prove the convergence based on this. Finally, we prove that our algorithm achieves at least 
\begin{math}
 O(\frac {1}{\sqrt{K}}+\frac {1}{K})
\end{math} convergence rate.

\subsubsection{Update Rules}
In CD-SGD, the global weights are updated by compressed gradients from all worker nodes, and there is no data loss in quantization. If the absolute value of a gradient is bigger than the threshold, it will be made the same value as the threshold and sent to server. The error between quantified gradients and original gradients is saved in residual buffer. The data saved in residual buffer cannot participate in the update until its absolute value exceeds the threshold. In every iteration, the global weights are updated on server nodes, and the local weights are updated on worker nodes. The update rules are shown in eq.\ref{eq10} and eq.\ref{eq11}, where $L_c(W_{i,k}^{loc};D_i)$ is the compressed gradient from the $i^{th}$ worker node, calculated with local weights $W_{i,k}^{loc}$ in the $k^{th}$ iteration.

\begin{equation}
 W_{k+1} = W_k - {\frac{\eta}{N}{\sum_{i=1}^N}{\triangledown}L_c(W_{i,k}^{loc};D_i)}
 \label{eq10}
\end{equation}

\begin{equation}
 W_{k+1}^{loc} = W_k - \eta{\triangledown}L(W_{i,k}^{loc};D_i)
 \label{eq11}
\end{equation}

\subsubsection{Convergence Rate Proof}
At first, we analyze the delay weight updates caused by compression will not affect the convergence of CD-SGD. Although the value of the uncompressed gradients is affected by the delayed updates, and it causes the weights over corrected sometimes. However, as the training goes on, the loss becomes smaller and smaller, then the affect can be ignored. Therefore, we can derive the relationship between the global weight and the uncompressed gradient according to the update rules and use the obtained conclusion as theorem 1. Then we can obtain the convergence limit based on theorem 1 and some reasonable assumptions.
% \theoremheaderfont{\bfseries\upshape}
% \theorembodyfont{\bfseries\upshape}

\noindent{\bfseries {\itshape Assumption 1.}}The data in the residual buffer participates in the gradient update every $n$ iterations on average, and the values of them are donated by $u$.

\noindent{\bfseries {\itshape Theorem 1.}}When the number of iteration is big enough that there is no extreme training loss, we have
\begin{equation}\label{eq12}
{W_k} \simeq {{W_{k-1}} - {\frac {\eta}{N}{\sum_{i=1}^N}{\triangledown}L(W_{i,k-1}^{loc};D_i)} +{\gamma}}
\end{equation}

%{\frac {\eta}{N}\noindent{\bfseries {\itshape Proof 1.}}
\begin{proof}
According to our assumption, the value of $u$ can be represented by the compressed gradient
\begin{math}
 {\triangledown}L_c(W_{i,j}^{loc};D_i)
\end{math}
and the uncompressed gradient 
\begin{math}
 {\triangledown}L(W_{i,j}^{loc};D_i).
\end{math}

\begin{displaymath}
\begin{split}
   u=&sign({\triangledown}L(W_{i,j-1}^{loc};D_i))|{\triangledown}L(W_{i,j-1}^{loc};D_i)-{\triangledown}L_c(W_{i,j-1}^{loc};D_i)|+\\ 
  & sign({\triangledown}L(W_{i,j}^{loc};D_i))|{\triangledown}L(W_{i,j}^{loc};D_i)-{\triangledown}L_c(W_{i,j}^{loc};D_i)| +\cdots+ \\
  & sign({\triangledown}L(W_{i,j+n-1}^{loc};D_i))|{\triangledown}L(W_{i,j+n-1}^{loc};D_i)-{\triangledown}L_c(W_{i,j+n-1}^{loc};D_i)|		 
  \end{split}
\end{displaymath}
With the help of $u$, we can get the relationship of $W_k$ and $W_{k-n}$:

\begin{equation}
\begin{split}
{W_k} = & {{W_{k-n}} - {\frac{\eta}{N}[{\sum_{j=k-n+1}^k}{\sum_{i=1}^N}{\triangledown}L(W_{i,j-1}^{loc};D_i)}-u ] }
  \end{split}
  \label{eq13}
\end{equation}
% \vspace{0.8em}
When using the eq.\ref{eq13} to recurse $W_k$, we can express $W_k$ in terms of $W_0$ and the sum of local compressed gradients, and $u$ is added every $n$ iterations.

\begin{equation}{
\begin{aligned}
{W_k} = & {{W_{k-1}} - {\frac {\eta}{N}{\sum_{i=1}^N}{\triangledown}L_c(W_{i,k-1}^{loc};D_i)} } \notag \\      =& {W_{k-2} - {\frac{\eta}{N}{\sum_{i=1}^N}{\triangledown}L_c(W_{i,k-2}^{loc};D_i)} -          {\frac{\eta}{N}{\sum_{i=1}^N}{\triangledown}L_c(W_{i,k-1}^{loc};D_i)}  } \notag \\   
  =& {W_0 - {\frac{\eta}{N}{\sum_{j=1}^k}{\sum_{i=1}^N}[{\triangledown}L(W_{i,j-1}^{loc};D_i)}-\frac{1}{n}u ] }  \notag \\ 
  = & W_0 - {\frac{\eta}{N}{\sum_{j=1}^{k/n}}{\sum_{i=1}^N}[{\triangledown}L(W_{i,j-1}^{loc};D_i) + {\triangledown}L(W_{i,j}^{loc};D_i) +{\cdots}} \\
  & {+ {\triangledown}L(W_{i,j+n-1}^{loc};D_i) -u ]}
  \end{aligned}}
\end{equation}
The value of u is always limited by threshold ${\alpha}$, so no matter {k/n} is an integer or not, we can get eq.\ref{eq12} from the above equation, in which $|{\gamma}|<\frac {\eta}{NK}{\alpha}$.
\end{proof} 
\noindent{\bfseries {\itshape Assumption 2.}} Besides theorem 1 and updated rules mentioned above, we need the following assumptions.
\begin{itemize}
\item For the optimum {\bfseries $W_*$} and any {\bfseries W},
\begin{math}
 ||W-W_*|| \leq R
\end{math}
\item For any {\bfseries W},
\begin{math}
 ||{\triangledown}(L(W)|| \leq G
\end{math}
\item For any 
\begin{math}
 j \in [K], i \in [N],
\end{math}
and W,
\begin{math}
 ||{\triangledown}L(W_{i,j};D_i)-{\triangledown}L(W)|| \leq \beta
\end{math}
\item If L has l-Lipschitz gradient, 
\begin{math}
 ||{\triangledown}L(u)-{\triangledown}L(v)|| \leq l||u-v||
\end{math}
holds for any u and v.
\end{itemize}

\noindent{\bfseries {\itshape Theorem 2.}} Let $\eta$ in eq.\ref{eq10} be the same as that in eq.\ref{eq11}. Under Assumption 2, we have

\begin{equation}
L(\frac {1}{K}{\sum_{k=1}^K}w_j) -L(w_*) \leq {\frac{{3\eta}(G+\beta+{\frac{\alpha}{NK}})^2}{2}}+{\frac{R\alpha}{NK}}+2lR\eta (G+\beta+{\frac{\alpha}{2NK}})
\end{equation}
\begin{proof} According to the update rules, we have
\begin{displaymath}
\begin{split}
  &\quad||W_{k+1}-{W_*}||^2\\
  &\leq ||W_{k}-{W_*}||^2-2\eta \big \langle {\frac {1}{N}{\sum_{i=1}^N}{\triangledown}L(W_{i,k}^{loc};D_i)} ,W_{k}-{W_*} \big \rangle+2{\frac{{\eta}R\alpha}{KN}} \\
  &\quad+ {\eta^2}(G+\beta+{\frac{\alpha}{NK}})^2
\end{split}
\end{displaymath}
Therefore,we have
\begin{displaymath}
\begin{split}
  &\big \langle {\frac{1}{N}{\sum_{i=1}^N}{\triangledown}L(W_{i,k}^{loc};D_i)},W_{k}-{W_*} \big \rangle\\ 
  &\leq {\frac{||W_{k}-{W_*}||^2}{2\eta}}-{\frac{||W_{k+1}-{W_*}||^2}{2\eta}}+{\frac{R\alpha}{NK}}+{\frac{{\eta}(G+\beta+{\frac{\alpha}{KN}})^2}{2}}
\end{split}
\end{displaymath}
Thus, we have
\begin{displaymath}
\begin{split}
  &\quad E\big \langle {\triangledown}L(W_k),W_{k}-{W_*} \big \rangle\\ 
   &\leq {\frac{||W_{k}-{W_*}||^2}{2\eta}}-{\frac{||W_{k+1}-{W_*}||^2}{2\eta}}+{\frac{R\alpha}{NK}}+{\frac{{\eta}(G+\beta+{\frac{\alpha}{NK}})^2}{2}}\\
   &\quad+2lR\eta (G+\beta+{\frac{\alpha}{2NK}})
\end{split}
\end{displaymath}
\end{proof}

\noindent{\bfseries {\itshape Corollary .}}Choose $\eta=\frac {R}{\sqrt{K}(G+\beta+\frac{\alpha}{NK})}$, we have

\begin{displaymath}
\begin{split}
L(\frac {1}{K}{\sum_{k=1}^K}w_j) -L(w_*)&\leq {\frac{3R(G+\beta+\frac{\alpha}{NK})}{2\sqrt{K}}}+{\frac{R\alpha}{NK}}+2\frac{lR }{\sqrt{K}}\\
 &\leq \frac {1}{\sqrt{K}}+\frac {1}{K}
  \end{split}
\end{displaymath}
That means CD-SGD has at least 
\begin{math}
 O(\frac {1}{\sqrt{K}}+\frac {1}{K})
\end{math}
convergence rate. K is the number of iterations.

\section{PERFORMANCE EVALUATION}
\label{5}
In this section, we evaluate performance of CD-SGD. At first we introduce the hardware configurations, datasets and neural networks used in our experiments. Secondly, we compare the convergence properties of CD-SGD with that of OD-SGD \cite{xu2020od}(a local update algorithm), S-SGD and BIT-SGD (2-bit quantification in MXNet). Besides, we analyze the influence of different $k$ values on CD-SGD. Finally, we assess the performance improvement of CD-SGD.
% \label{shiyan}
\subsection{Experimental Settings}
% {\itshape }
{\bfseries Cluster Configuration: } We conduct experiments on two different 4-node clusters connected with 56Gbps InfiniBand. One of cluster installs 2 K80 (dual GPUs) Tesla GPUs on each node, and the other cluster is equipped with 4 V100 Tesla GPUs on each node. The K80 GPU cluster provides Red Hat 4.8.3, CUDA 8.0 and CUDNN 6.0, while the V100 GPU cluster nodes are installed with Centos 7.6, CUDA 10.0 and CUDNN 7.4.1. We propose CD-SGD on MXNet at version 1.4.1.

\noindent {\bfseries Datasets: } We do the experiment on MNIST, CIFAR-10, and ImageNet ILSVRC2012 datasets. MNIST \cite{mnist} is a handwritten digits database with a training set of 60,000 examples and a test set of 10,000 examples. CIFAR-10 \cite{krizhevsky2009learning} is a labeled subset of tiny images dataset which consists of 60000 32x32 color images(50000 training images, 10000 test images) in 10 classes. ImageNet ILSVRC2012 \cite{deng2009imagenet} is a subset of ImageNet and contains 1.2 million pictures in 1000 categories.

\begin{figure}[h]
  \centering
  \begin{subfigure}[h]{0.5\linewidth}
    \includegraphics[width=\linewidth]{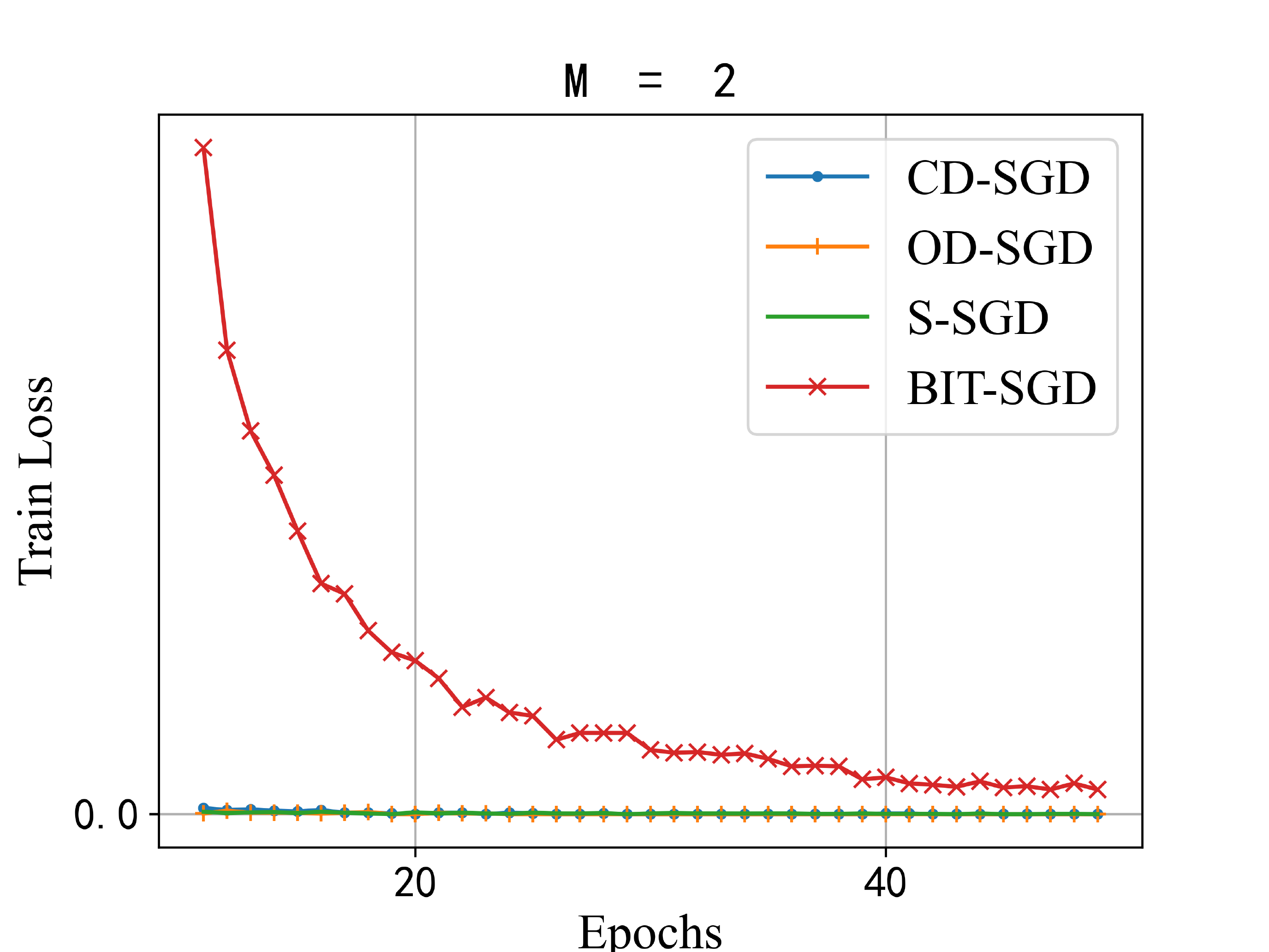}
    \caption{Training loss (M=2)}
    \label{mn2l}
  \end{subfigure}
  ~
  \begin{subfigure}[h]{0.5\linewidth}
    \includegraphics[width=\linewidth]{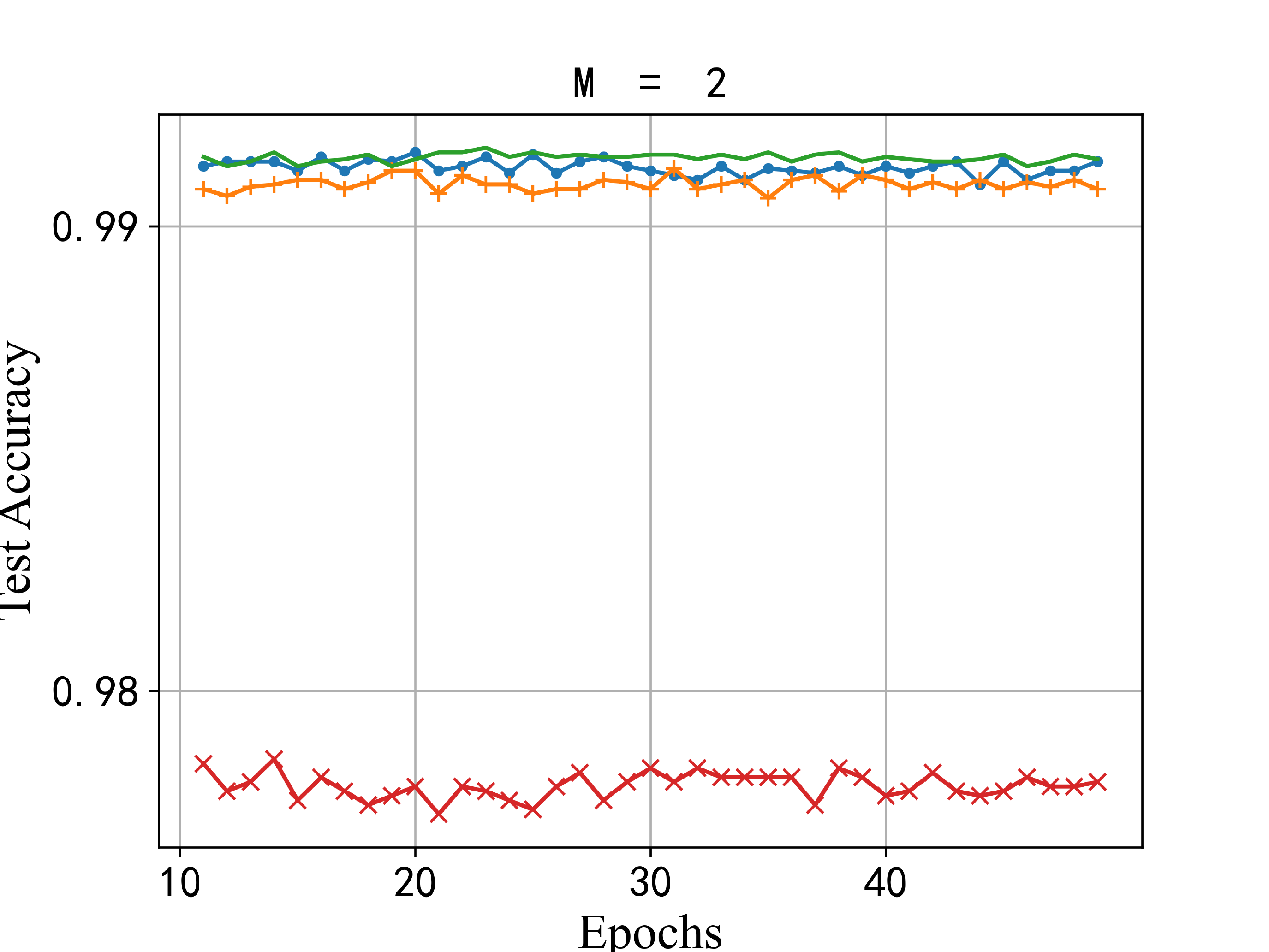}
    \caption{Test accuracy (M=2)}
    \label{mn2t}
  \end{subfigure}
  ~

  \begin{subfigure}[h]{0.5\linewidth}
    \includegraphics[width=\linewidth]{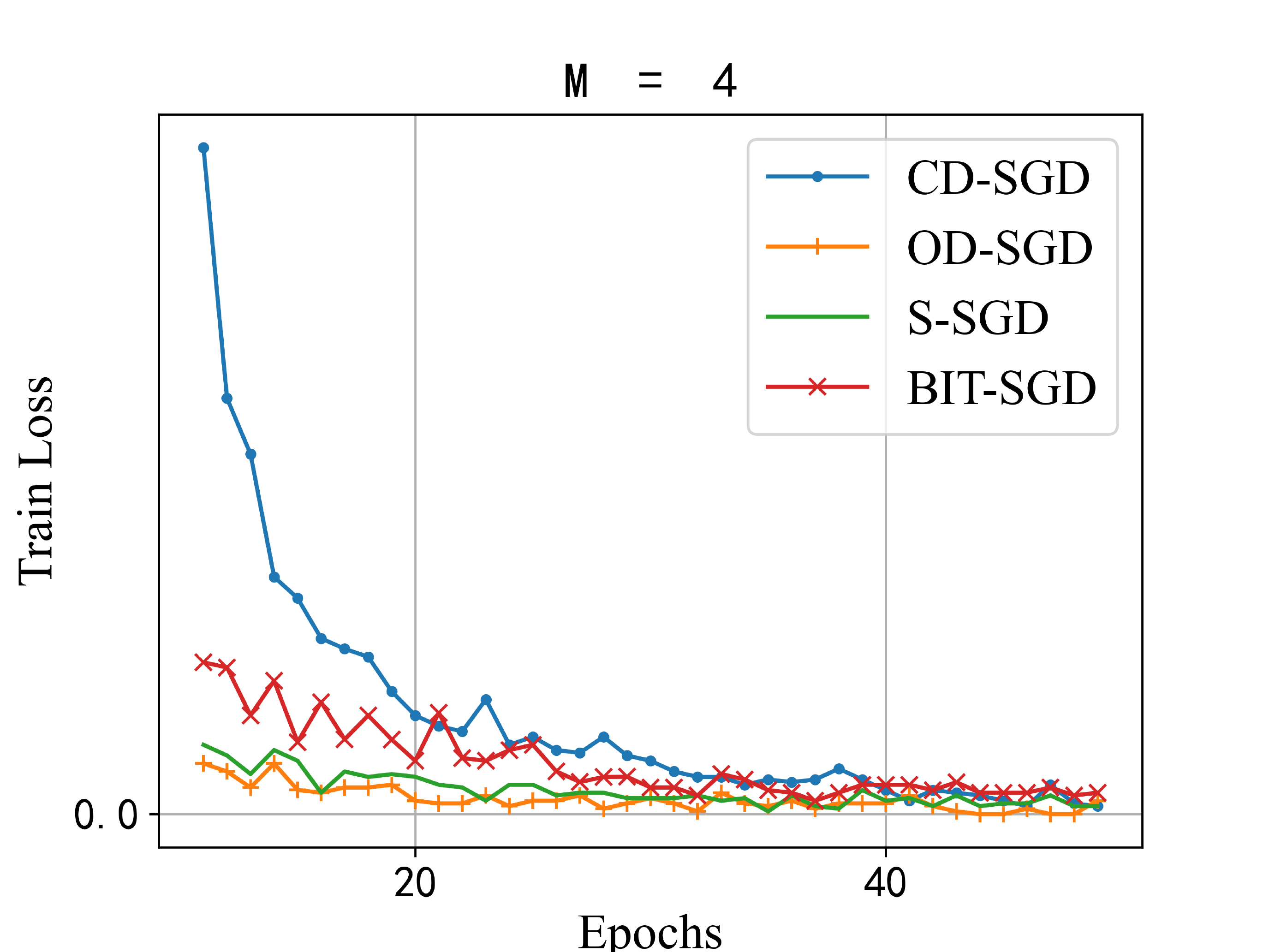}
    \caption{Training loss (M=4)}
    \label{mn4l}
  \end{subfigure}
  ~
  \begin{subfigure}[h]{0.5\linewidth}
    \includegraphics[width=\linewidth]{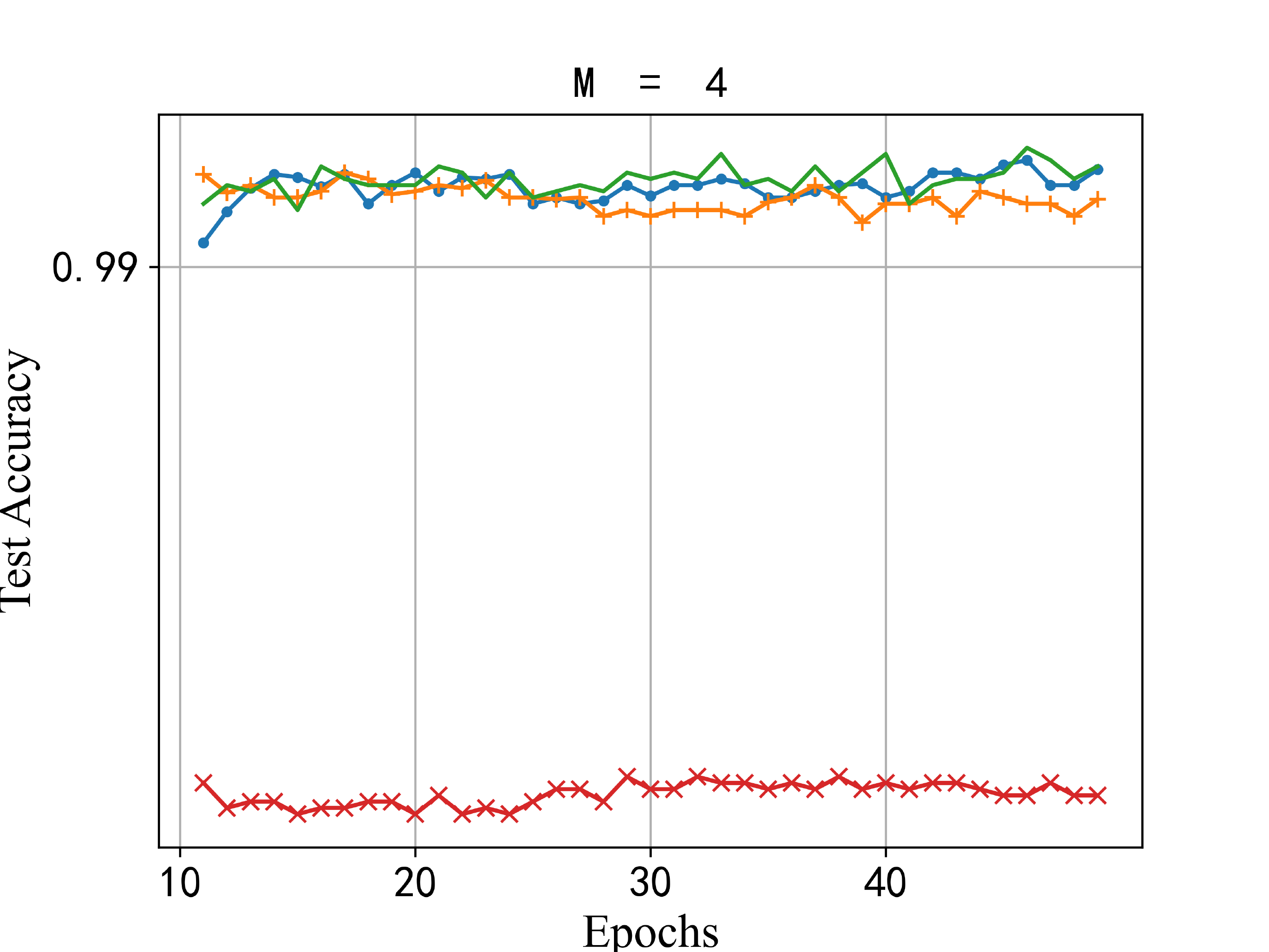}
    \caption{Test accuracy (M=4)}
    \label{mn4t}
  \end{subfigure}
  \caption{ Learning curve of Lenet-5 on MNIST. M represents the number of worker nodes}
  \label{mn4}
\end{figure}

\subsection{Convergence Performance}

%给一个cifar10的精度图一个image的精度图
In this subsection, we validate the convergence accuracy of CD-SGD on K80 cluster and V100 cluster. 
In order to observe the performance of CD-SGD in different cases, we set three groups of experiment: (1) Training Lenet-5 model with MNIST dataset on K80 GPU. (2) Training Inception-bn with CIFAR-10 dataset on K80 GPU. (3) Training ResNet-50 with ImageNet ILSVRC2012 dataset on V100 GPU.

When training Lenet-5, the global lr for all algorithms is 0.1 and the local lr of CD-SGD and OD-SGD is 0.4. Besides, the threshold in CD-SGD and BIT-SGD is 0.5 and the batch size on every GPU is 32. The value of k-step in CD-SGD is 2. Fig.\ref{mn4} is the test accuracy and training loss of Lenet-5 on MNIST. When training with 2 workers, the training loss of BIT-SGD performs worse than that of the other three approaches obviously, and its test accuracy is always lower than 99\%. However, CD-SGD solves the accuracy problem of BIT-SGD, and its convergence performance (99.14\%) is very close to S-SGD (99.15\%), even higher than OD-SGD (99.12\%). When training with 4 worker nodes, training loss of the four algorithms has some changes. The train accuracy of CD-SGD is lower than BIT-SGD, but the convergence accuracy of it is still higher than that of BIT-SGD and OD-SGD. We attribute the changes to the fact that Lenet model is too simple to be affected by the randomness of the mini-batch data set.

\begin{figure}
  \centering
  \begin{subfigure}[h]{0.5\linewidth}
    \includegraphics[width=\linewidth]{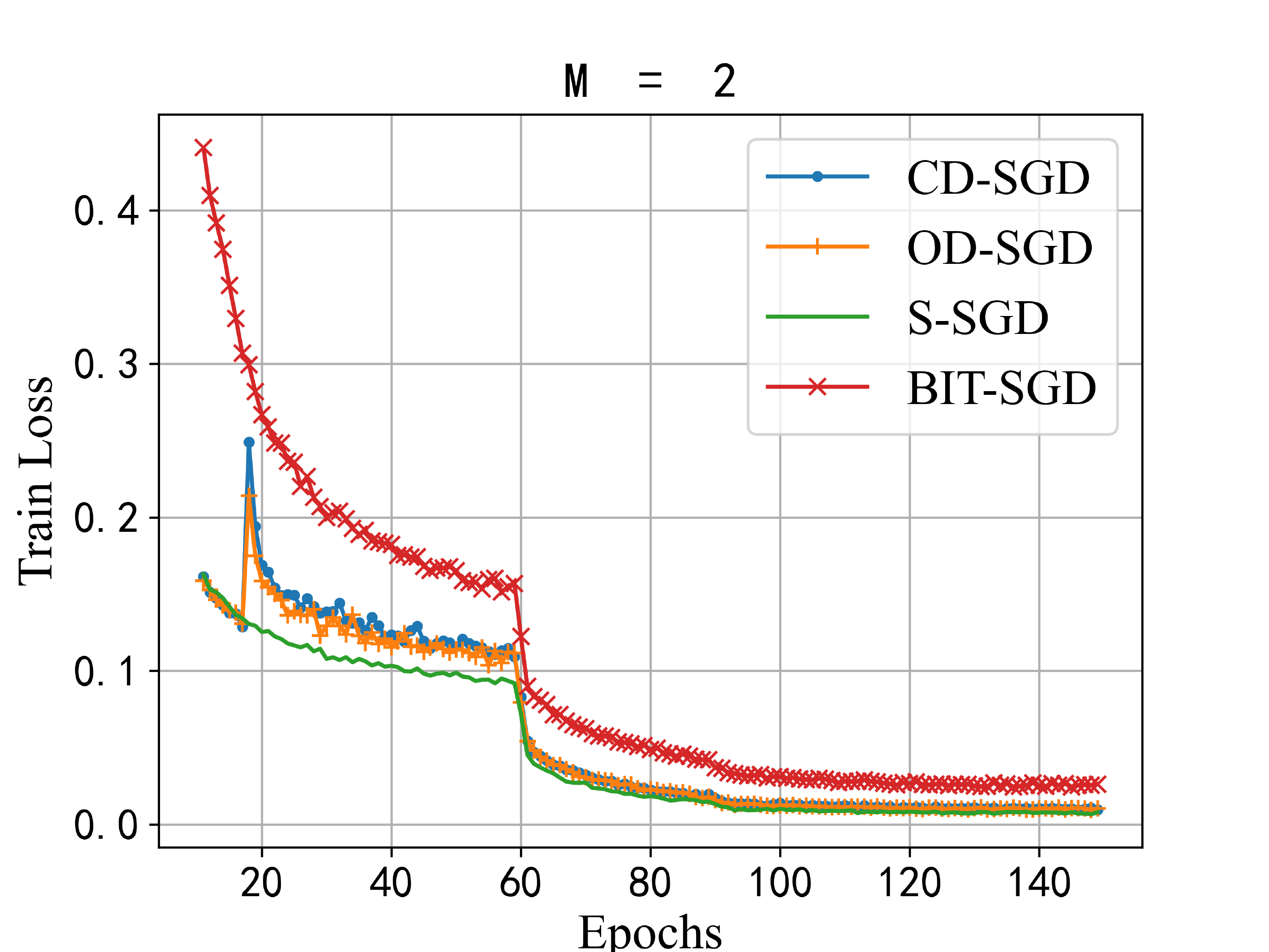}
    \caption{Training loss (M=2)}
    \label{bn2l}
  \end{subfigure}
  ~ 
  \begin{subfigure}[h]{0.5\linewidth}
    \includegraphics[width=\linewidth]{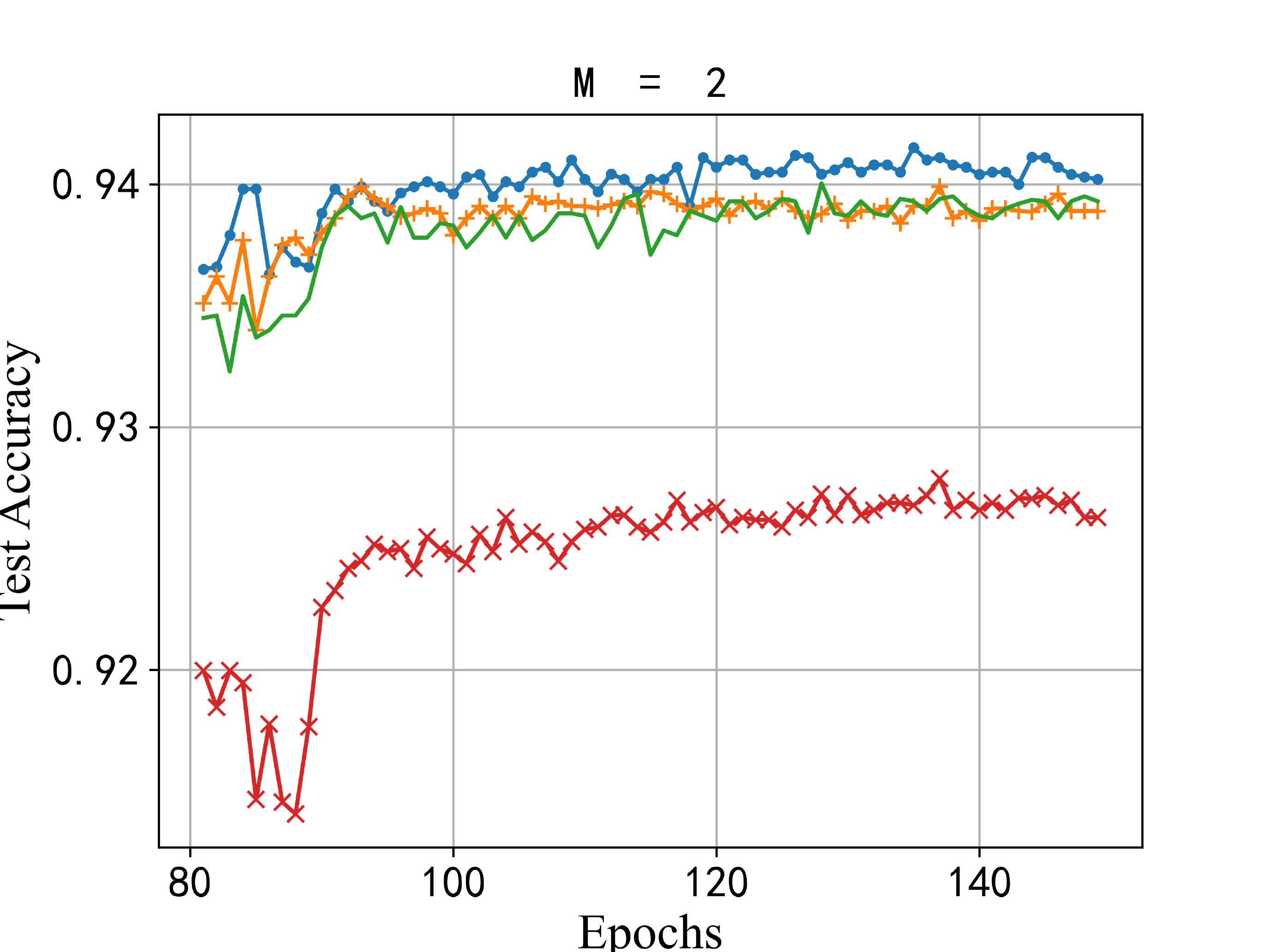}
    \caption{Test accuracy (M=2)}
    \label{bn2t}
  \end{subfigure}
  ~

  \begin{subfigure}[h]{0.5\linewidth}
    \includegraphics[width=\linewidth]{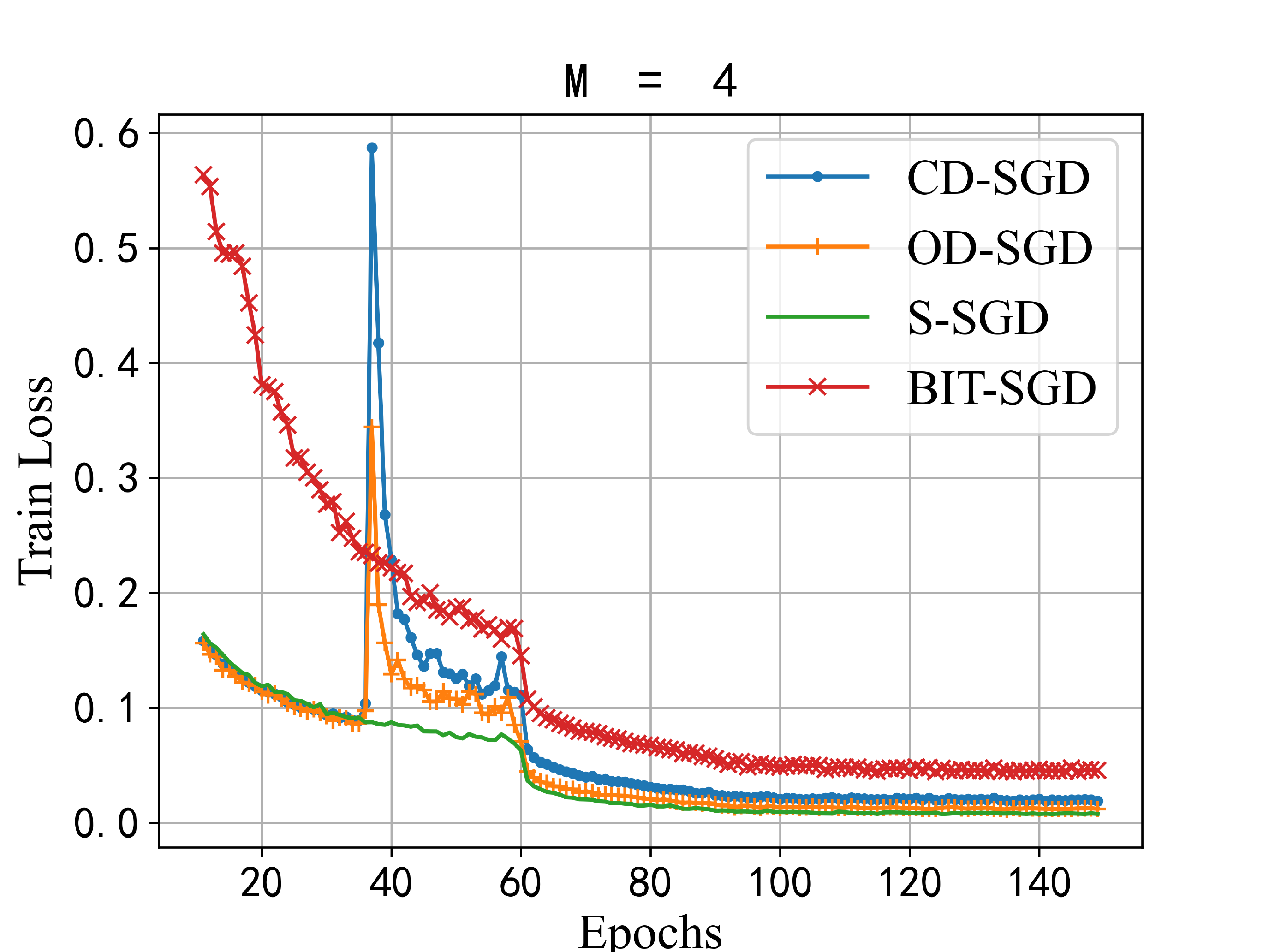}
    \caption{Training loss (M=4)}
    \label{bn4l}
  \end{subfigure}
  ~ 
  \begin{subfigure}[h]{0.5\linewidth}
    \includegraphics[width=\linewidth]{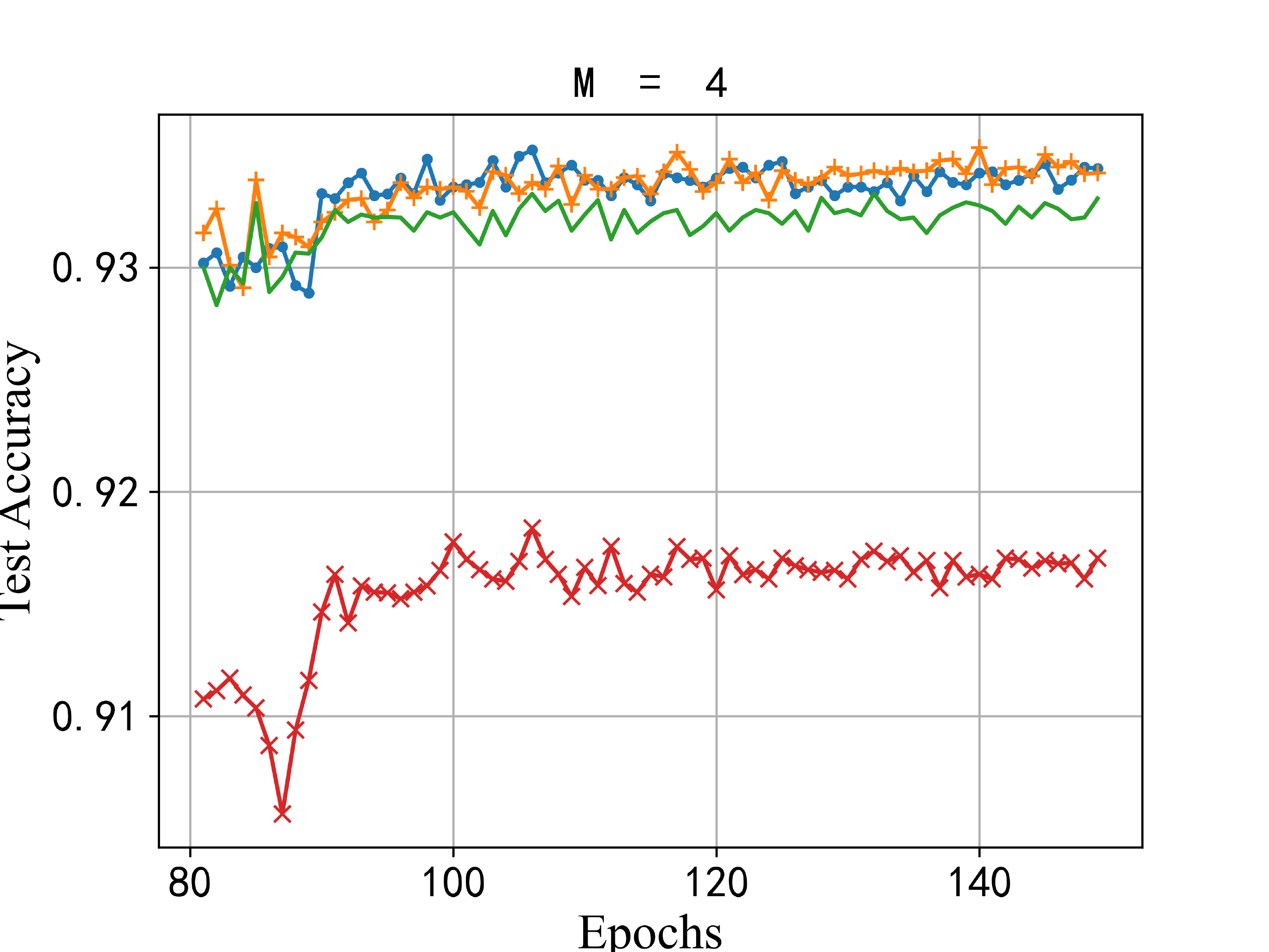}
    \caption{Test accuracy (M=4)}
    \label{bn4t}
  \end{subfigure}
  \caption{ Learning curve of Inception-bn on CIFAR-10. M represents the number of worker nodes}
  \label{bn4}
\end{figure}

When training Inception-bn, the global lr for all algorithms is 0.4 and the local lr of CD-SGD and OD-SGD is 0.05, and the other super parameters are unchanged. Fig.\ref{bn4} shows the performance of various algorithms on CIFAR-10 dataset. When training with 2 workers, although the training loss of CD-SGD is higher than that of OD-SGD and S-SGD, its test accuracy is the best among these four algorithms. The top-1 accuracy of CD-SGD, OD-SGD, S-SGD and BIT-SGD are 94.15\%, 93.99\%, 94.00\% and 92.69\% respectively. When training with 4 workers, CD-SGD still performs best. The top-1 accuracy shown in Fig.\ref{bn4} are 93.50\%, 93.50\%, 93.30\% and 91.83\%. It can be observed that there are large fluctuations in Fig.\ref{bn4}c during the training process, which is caused by the switch from the warm-up phase to the formal training state.

\begin{figure}
  \centering
  \begin{subfigure}[t]{0.5\linewidth}
    \includegraphics[width=\linewidth]{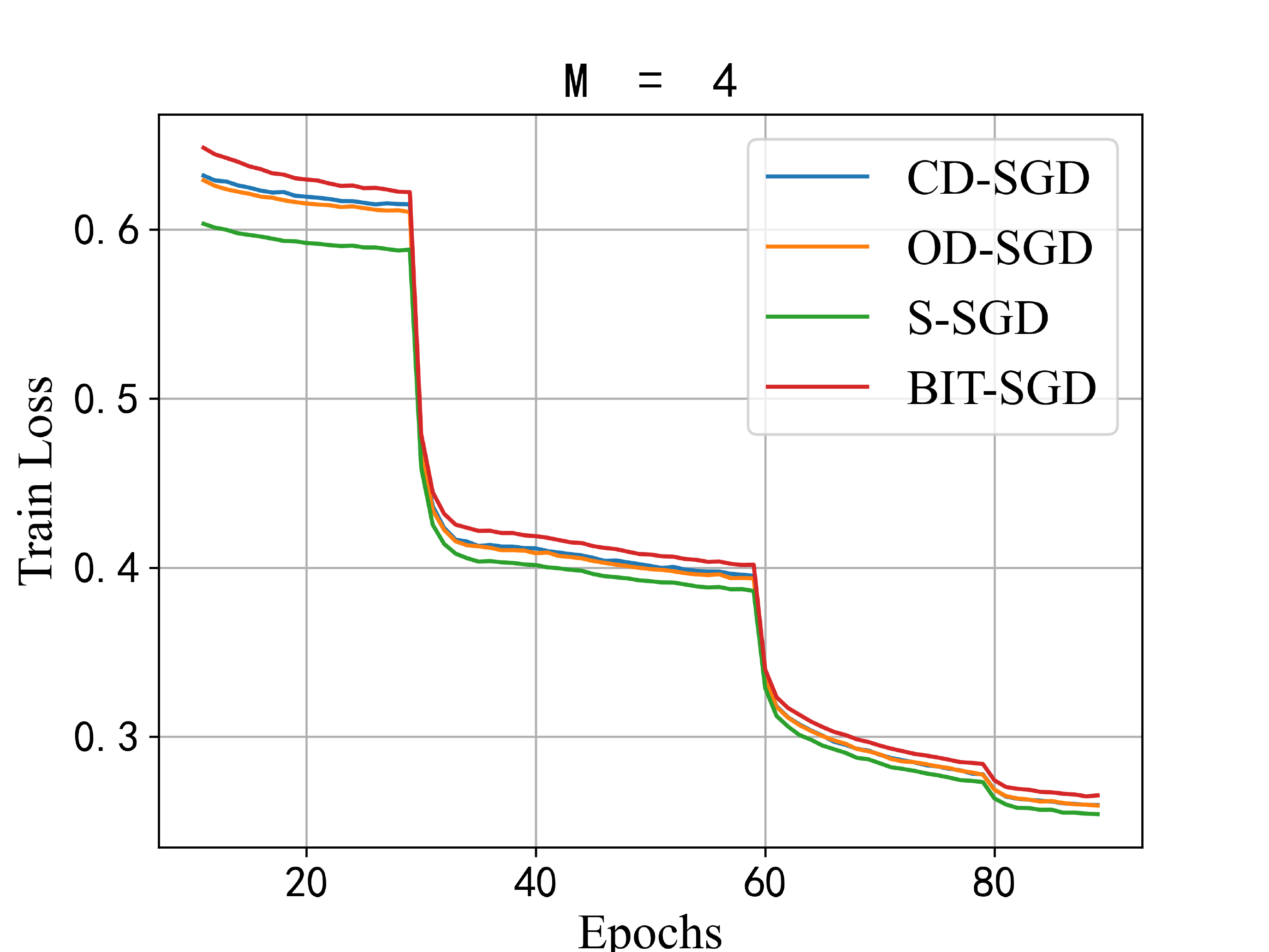}
    \label{r5l}
  \end{subfigure}
  ~ 
  \begin{subfigure}[t]{0.5\linewidth}
    \includegraphics[width=\linewidth]{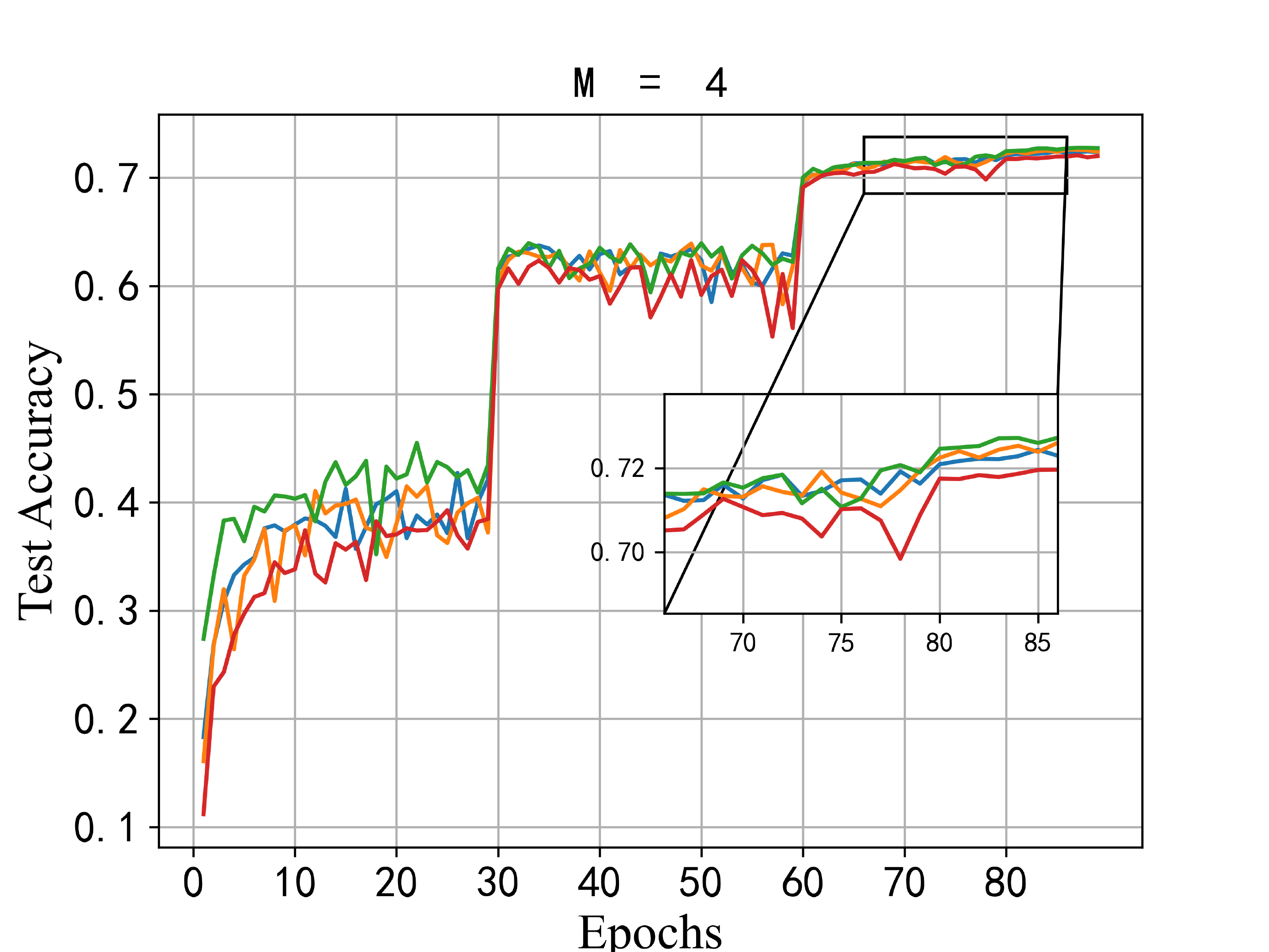}
    \label{r5t}
  \end{subfigure}
  \caption{ Learning curve of ResNet-50 on ImageNet with 4 workers}
  \label{r5}
\end{figure}

When training ResNet-50, the local lr of CD-SGD and OD-SGD is changed to 0.1, and the learning rate is adjusted at the 30th, 60th and 80th epoch. As shown in Fig.\ref{r5}, the training loss and test accuracy of BIT-SGD are always worse than others. CD-SGD obtains almost the same test accuracy as OD-SGD, but a little inferior than S-SGD. The top-1 accuracy of these four algorithms are 72.4\%, 72.6\%, 72.7\%, 72.0\%. Therefore, we could a conclusion that when a complex model is trained on ImageNet ILSVRC2012 dataset, the convergence accuracy of CD-SGD, OD-SGD, S-SGD and BIT-SGD is roughly the same, and the main difference between them is training speed. In this experiment, the average epoch time cost of CD-SGD is 41\% less than that of BIT-SGD.

\subsection{K-step Sensitivity Analysis}

\begin{figure}
  \centering
  \begin{subfigure}[h]{0.5\linewidth}
    \includegraphics[width=\linewidth]{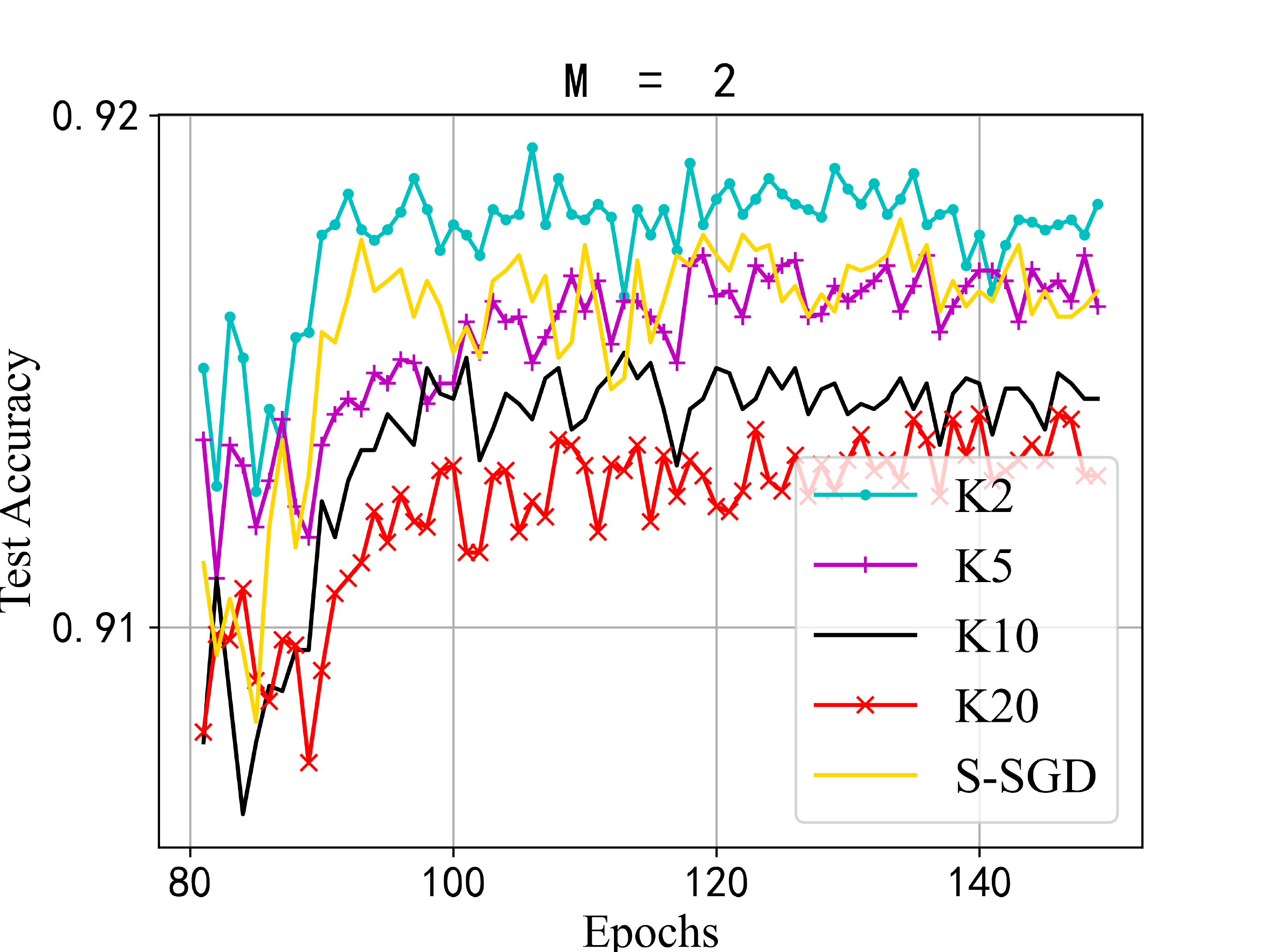}
    \caption{K-step Sensitivity Analysis on 2 nodes}
    \label{2mg}
  \end{subfigure}
  ~ %add desired spacing between images, e. g. ~, \quad, \qquad, \hfill etc. 
   %(or a blank line to force the subfigure onto a new line)
  \begin{subfigure}[h]{0.5\linewidth}
    \includegraphics[width=\linewidth]{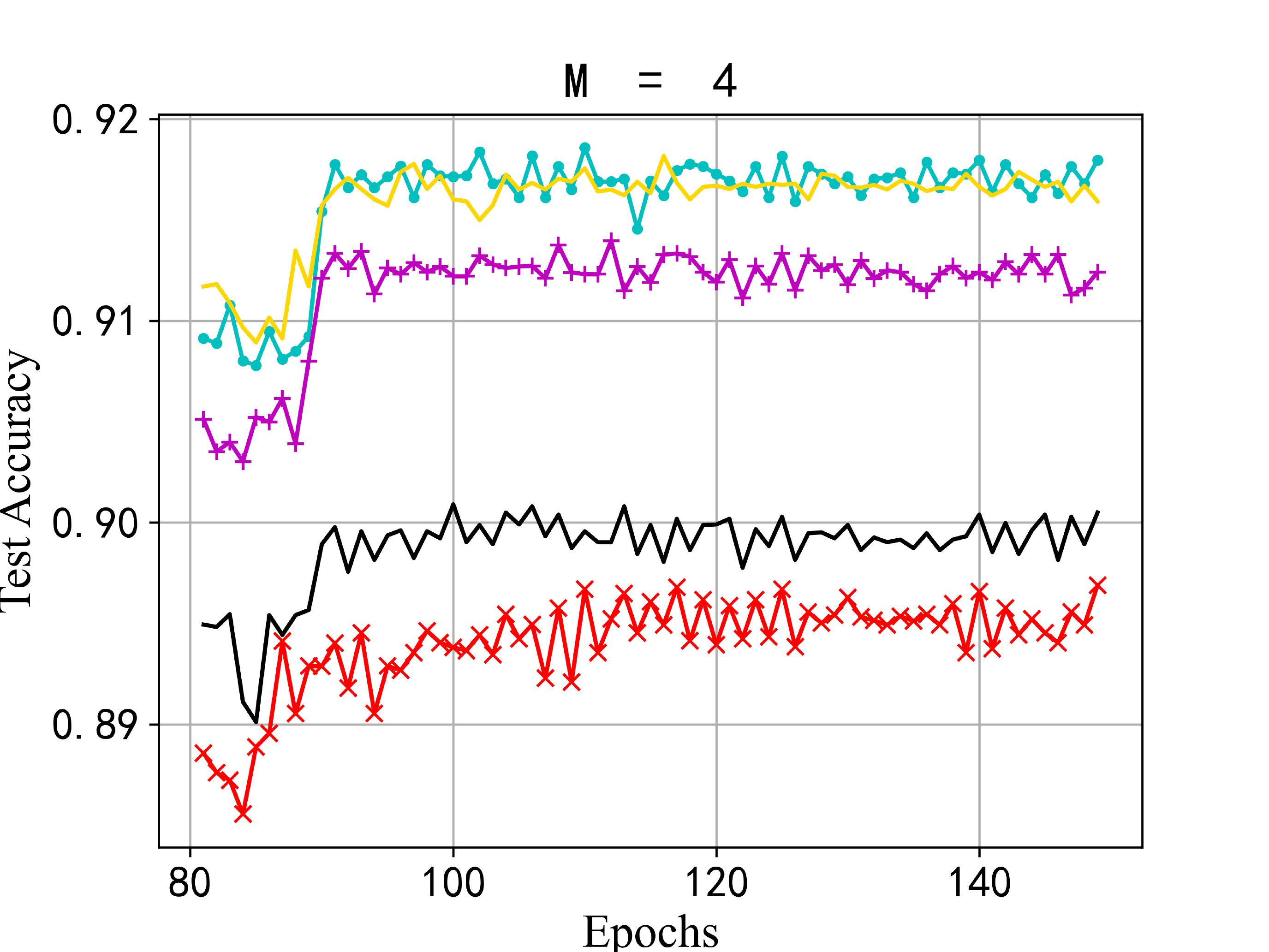}
    \caption{K-step Sensitivity Analysis on 4 nodes}
    \label{4mg}
  \end{subfigure}
  \caption{ Test accuracy of ResNet-20 (CIFAR-10, with data augmentation)
under different values of k.}
\label{mg}
\end{figure}

In section \ref{4}, we introduce the importance of k-step correction. To analyze the influence of $k$ on the accuracy improvement, we train ResNet-20 on CIFAR-10 dataset with 2 and 4 worker nodes respectively. Fig.\ref{mg} illustrates the accuracy of CD-SGD with different $k$ values, from which we can get the following observations: Firstly, CD-SGD always achieves the best convergence accuracy when the value of $k$ is 2, which is better than S-SGD. And the accuracy of $k_5$ is close to that of S-SGD. Besides, the convergence accuracy decreases as $k$ increases, this phenomenon is obvious when more nodes are used for training. When $k$ becomes infinite, CD-SGD can be regarded as an algorithm without k-step correction. In this case, its accuracy is close to that of BIT-SGD. For example, the top-1 convergence accuracy of $k_{20}$ in Fig.\ref{mg}b is 89.68\%, which is close to that of BIT-SGD (88.81\%). Table \ref{t2} records the average epoch wall-clock time of various algorithm, which shows that $k$ has no effect on training speed because computing is the bottleneck on K80. In this case, the speed advantage of CD-SGD comes from the parallelism of computation and communication. Based on the experimental results above, we have the following conclusion: It is effective to deal with the decrease of convergence accuracy by k-step correction, especially when $k$ is 2, CD-SGD can achieve better convergence accuracy than S-SGD. Besides, it is usually appropriate to choose k equal to 5 for practical application, when CD-SGD can achieve good accuracy and performance. If better accuracy is required, the value of k can be reduced, and if higher performance is pursued, the value of k can be increased. 
%需要补充v100的速度 这样才能分析k对速度的影响与什么有关

\begin{table}[h] 
 \caption{The average epoch wall-clock time of ResNet-20 on CIFAR-10 (in seconds)}
 \resizebox{\linewidth}{!}{
  \begin{threeparttable}
\begin{tabular}{ | l | l | l | l | l | l | l | l | }
\hline
Model  & SSGD & BIT-SGD & $k_2$  & $k_5$  & $k_{10}$ & $k_{20}$ \\
\hline
Resnet20(4nodes) & 2.24 & 2.22  & 1.79 & 1.78 & 1.78 & 1.76 \\
\hline
Resnet20(2nodes) & 4.32 & 3.61  & 3.48 & 3.44 & 3.46 & 3.44 \\
\hline
\end{tabular}
   \begin{tablenotes}
    \footnotesize
    %\item The quick brown fox jumps over the lazy dog. The quick brown fox jumps over the lazy dog. and $"k_0"$ means training without compression,th
   
\item[1] $"k_2"$ means the value of $k$ is 2, and so on.
   \end{tablenotes}
  \end{threeparttable}}
  \label{t2}
\end{table}

\subsection{Speedup on Different Models}

\begin{figure}
  \centering
  \begin{subfigure}[h]{0.5\linewidth}
    \label{fig:k80-speed}
    \includegraphics[width=\linewidth]{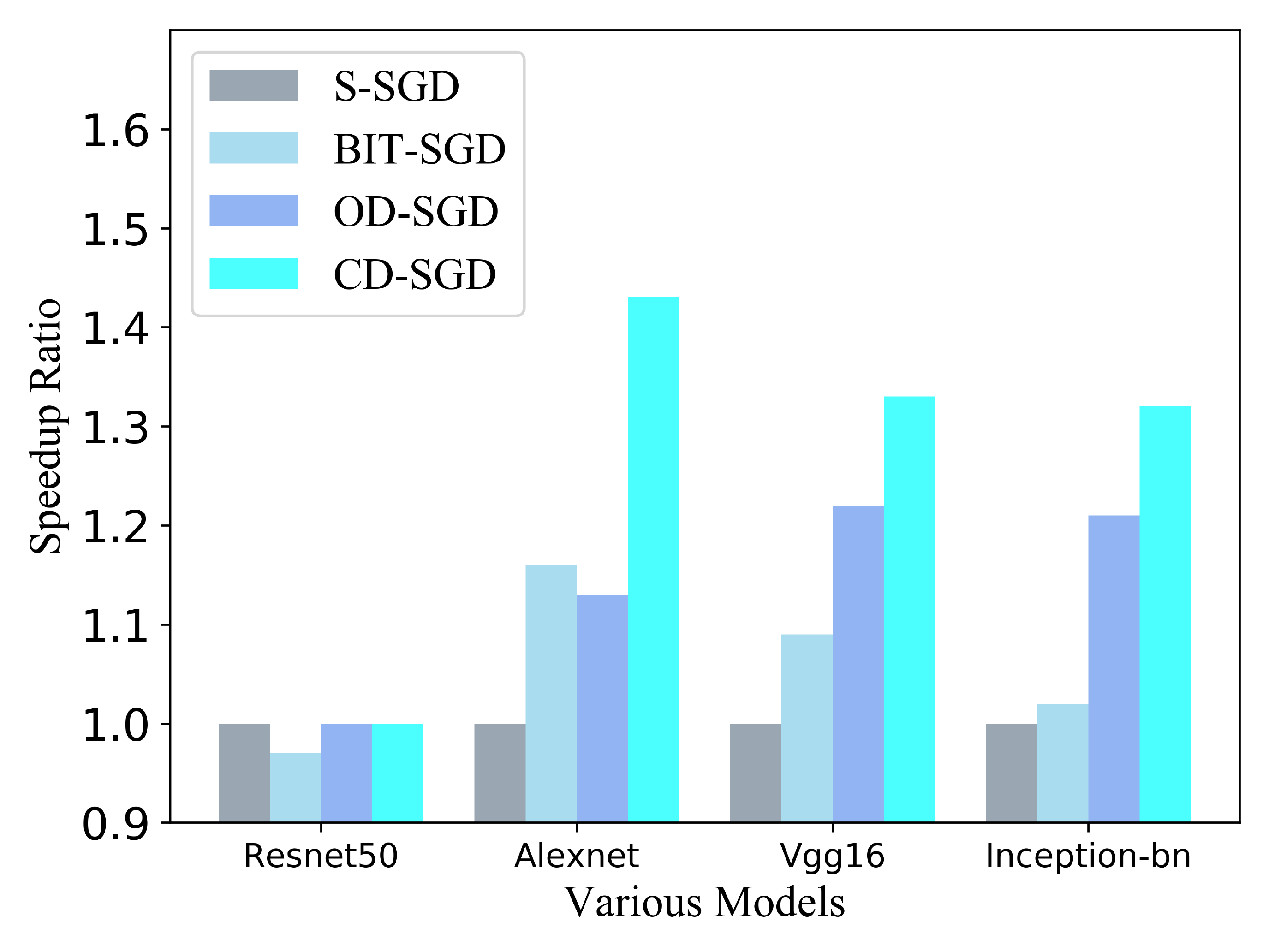}
        \caption{batch size 32 per GPU on K80}
  \end{subfigure}
  ~
  \begin{subfigure}[h]{0.5\linewidth}
    \includegraphics[width=\linewidth]{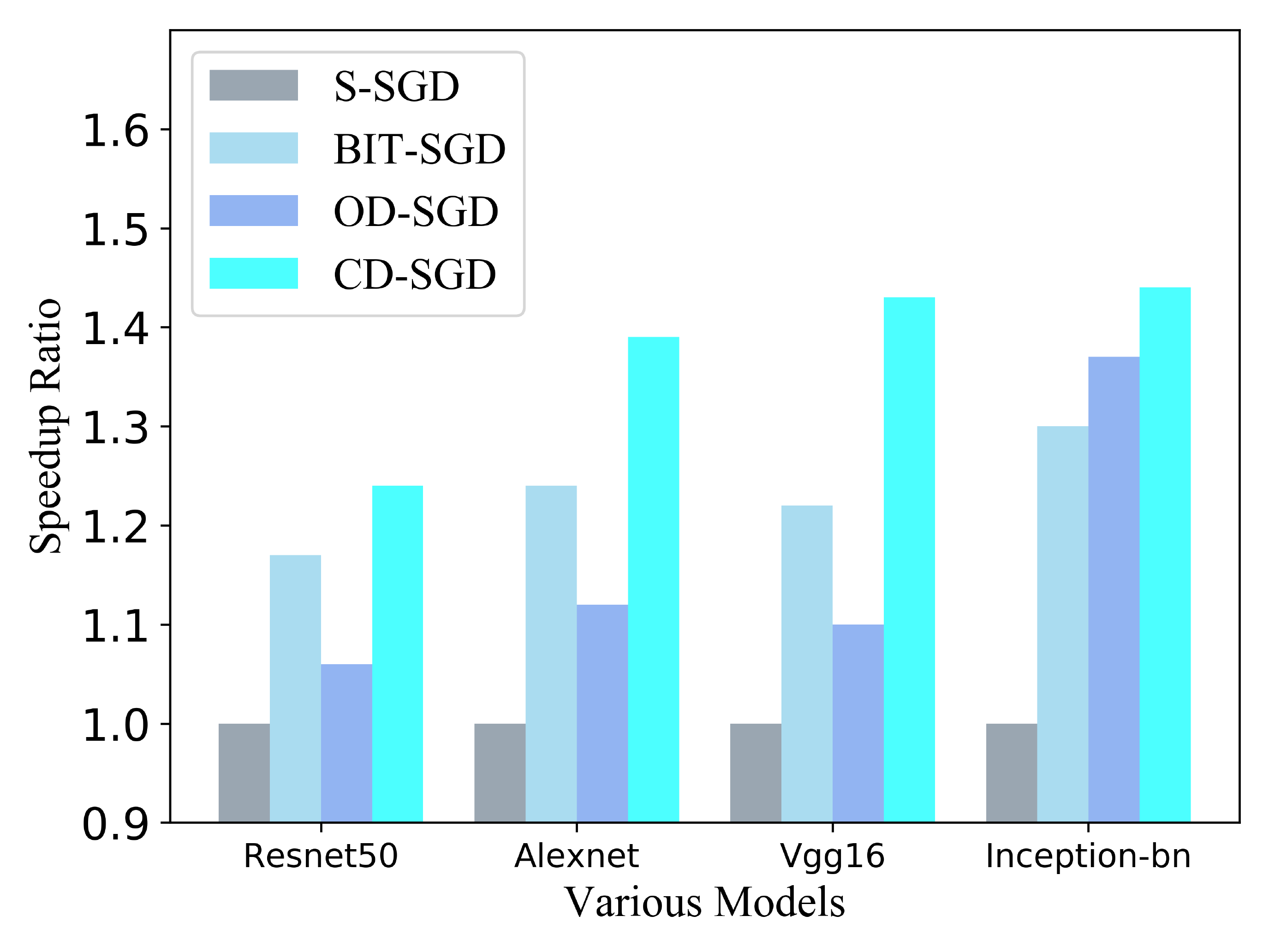}
    \caption{batch size 32 per GPU on V100}
    \label{v32}
  \end{subfigure}
  ~
  
  \begin{subfigure}[h]{0.5\linewidth}
    \includegraphics[width=\linewidth]{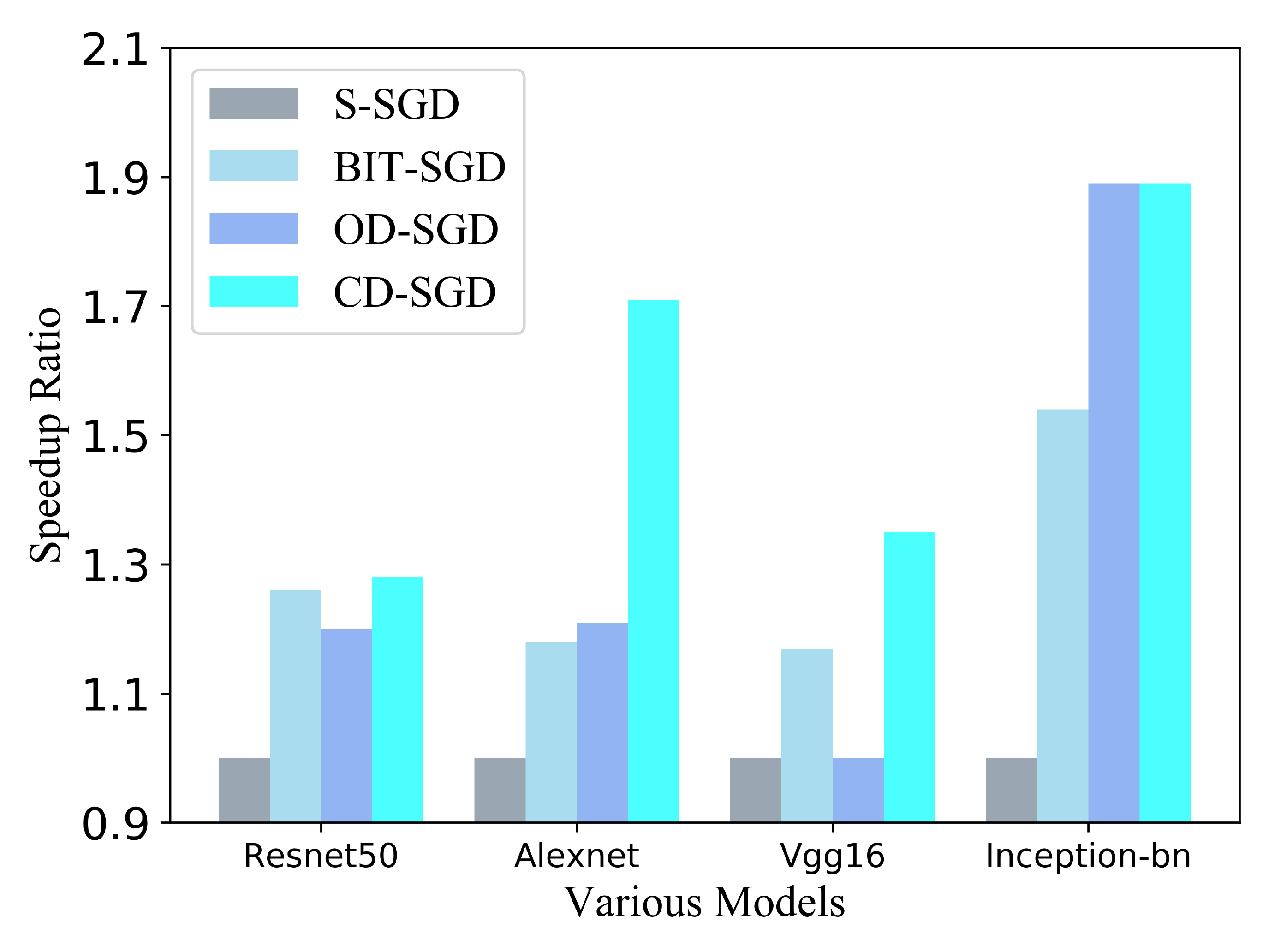}
        \caption{batch size 64 per GPU on V100}
        \label{v64}
  \end{subfigure}
  ~
  \begin{subfigure}[h]{0.5\linewidth}
    \includegraphics[width=\linewidth]{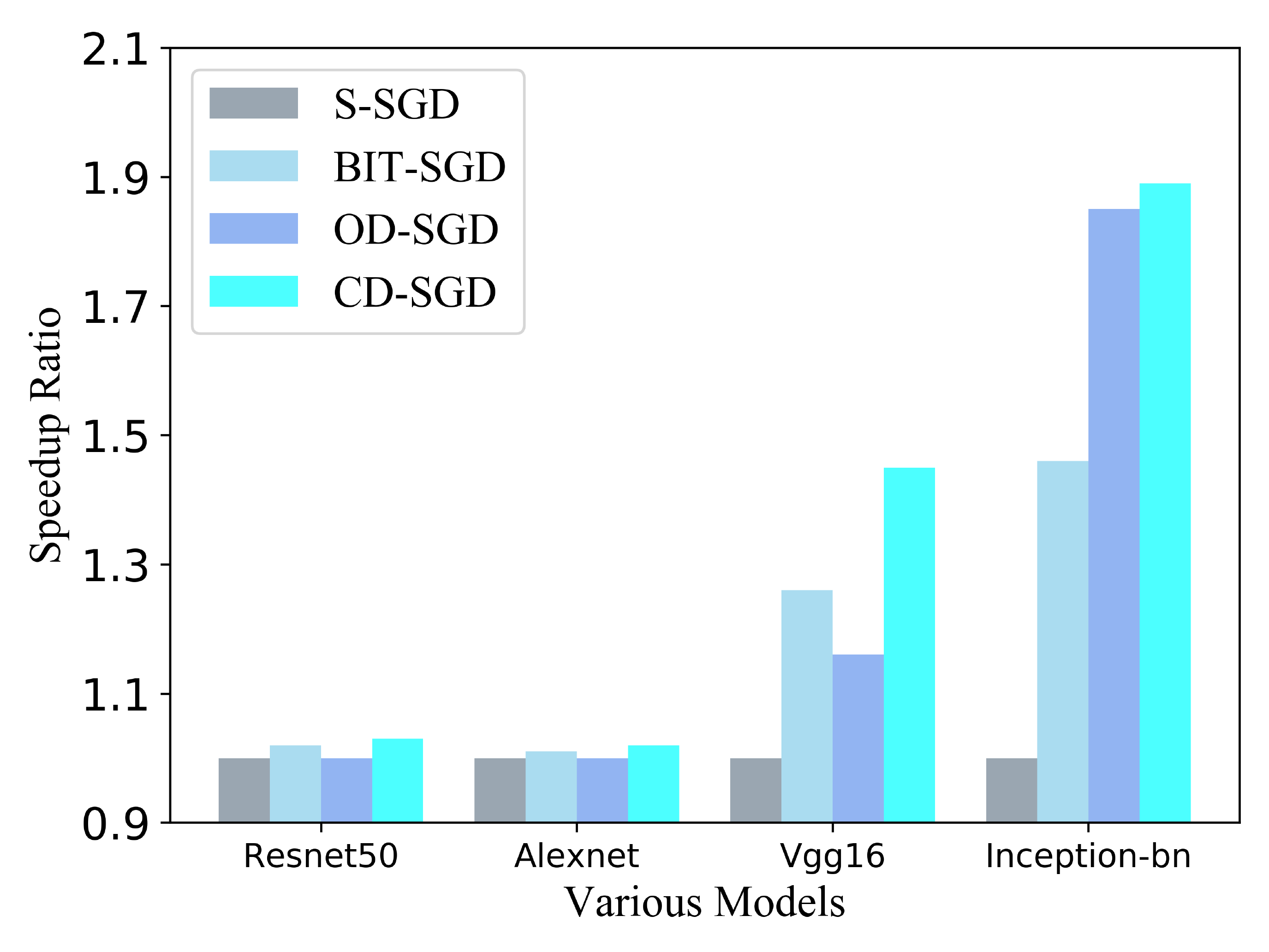}
        \caption{batch size 128 per GPU on V100}
        \label{v128}
  \end{subfigure}
  \caption{The speedup ratio of various models training with 4 worker nodes on K80 cluster and V100 cluster. The value of $k$ in CD-SGD is 5.}
  \label{fig:speed}
\end{figure}

We test the speed of various algorithms with different batch size training on K80 and V100 separately. In order to observe the speed comparison of each algorithm, we take the speed of S-SGD as the baseline. When training ResNet-50 on K80 in Fig.\ref{fig:speed}a, CD-SGD get the same training speed as OD-SGD, we attribute this phenomenon to the limited computing power of K80, which leads to the bottleneck of computation. And at the end of section \ref{II}, we have a analysis for this phenomenon, it is consistent with case 1 of 
\begin{math}
 T_{s}^{loc}
\end{math}
, and the difference between BIT-SGD and CD-SGD results from the extra compression cost. Besides, we can notice that BIT-SGD performs worse than OD-SGD when training Vgg16 and Inception-bn, which differs from Alexnet. It means that in addition to covering up the compression overhead and reducing the communication time, it is necessary to use parallel mechanism to increase the overlap of computation and communication. In the speed test experiment, the k-step of CD-SGD is 5. The speedup ratio of CD-SGD on models shown in Fig.\ref{fig:speed}a are 0\%, 43\%, 33\%, 32\%. The V100 GPU has more computing power, which can complete the computation task in less time. So, BIT-SGD performs better than OD-SGD when training most models in Fig.\ref{fig:speed}b because the limited computation cost cannot cover up communication time completely. However, BIT-SGD is slower than OD-SGD when training Inception-bn, because Inception-bn has many computation layers which leads to huge computation cost. The speedup ratio of CD-SGD on models shown in Fig.\ref{fig:speed}b are 24\%, 43\%, 39\%, 44\%. As the batch size becomes bigger, the speed of BIT-SGD is close to that of OD-SGD, and the acceleration effect of CD-SGD is weaker. The reason for this phenomenon is the larger batch size training brings greater computational pressure. In this case, the computation becomes the bottleneck of training. The speedup ratio of CD-SGD on models shown in Fig.\ref{fig:speed}c and Fig.\ref{fig:speed}d are 28\%, 35\%, 71\%, 89\% and 3\%, 45\%, 2\%, 89\%. Compared with BIT-SGD, CD-SGD can speed up by 3\% to 45\%.

% \begin{figure}[h]
%  \centering
%  \includegraphics[width=\linewidth]{resultk80.pdf}
%  \caption{ The speedup ratio of various models training with 4 worker nodes on K80 cluster. We normalized the training speed of 2-bit SGD as the baseline.
% }
% \end{figure}

\section{RELATED WORK}
\label{6}
In this section we introduce some related work of distributed communication optimization. We classify them into three categories.

{\bfseries Reducing Communication Overhead: }
The related work of reducing communication overhead is divided into gradient compression and more efficient communication mechanism. QSGD \cite{alistarh2017qsgd} and Terngrad \cite{ wen2017terngrad} provide gradient quantization methods and give corresponding proof of convergence. DGC \cite{lin2017deep} and SparCML \cite{renggli2019sparcml} transfers top-k gradients each time to decrease the communication traffic. Network Pruning \cite{2015Learning} reduces the model size to cut down the amount of network communication and computation and then Eager Pruning \cite{liu2018rethinking} further develops in this aspect.

{\bfseries Overlapping Communication and Computation: }
The optimization of synchronization training is represented by parallelization \cite{wang2019scalable,xu2020od}, hierarchical communication priority scheduling \cite{hashemi2018tictac,peng2019generic} and design of new topology protocol \cite{dong2020eflops,zhang2017poseidon}. Besides, LAGS-SGD \cite{shi2019layer} and OMGS-SGD \cite{shi2020communication} combine pipeline and gradient sparsity to hide the gradient overhead and reduce the communication time of a single iteration. These works are similar to CD-SGD, but CD-SGD is more focused on hiding the extra overhead of compression and improving the parallelism of computation and communication, and they are more concerned about improving the sparsification effect.

{\bfseries Reducing Communication Times: }
 Communication times can be reduced by increasing the batch-size, performing communication operations periodically instead of every iteration and only sending data to a part of worker nodes. In the early days, Facebook \cite{goyal2017accurate} could use 8K batch-size to train, and then LARS \cite{you2017large} further expands batch-size to 32K, and so on \cite{you2019larges,sun2019optimizing}. Use Local SGD \cite{lin2018don} speeds up training by communicating once during several iterations. There are similar works \cite{2018On,stich2018local,haddadpour2019local}. In Eager-SGD \cite{2020Taming}, it is considered that more than half of all worker nodes provide useful information is ok, and \cite{xie2016lighter,assran2019stochastic} optimize distributed training by updating weights with data from parts of the whole nodes.

\section{CONCLUSION}
\label{7}
In this paper, we first introduce the method of distributed communication optimization from the level of system optimization and algorithm optimization. Then we discuss the feasibility of combining system optimization with algorithm optimization and propose the current challenges. On this basis, we formulate the extra compression computation time cost and the decrease of convergence accuracy caused by compression as an optimization problem. And we provide an efficient optimal solution with theoretical guarantees to solve it. Via optimizing gradient compression with parallel mechanism and k-step correction method, we propose CD-SGD algorithm for distributed DNN training acceleration. We implement CD-SGD on MXNet framework and evaluate its performance on two clusters using various models. Experimental results showed that CD-SGD can speed up 2-bit quantization by 30\% and achieve slightly better convergence accuracy than S-SGD on a 16-GPU K80 cluster. And it performs even better on V100 cluster.

For the future work, we plan to evaluate CD-SGD on larger computer clusters with low bandwidth environment and it is worthy to explore efficient gradient sparsification algorithms to further improve the training efficiency of CD-SGD.

%%
%% The acknowledgments section is defined using the "acks" environment
%% (and NOT an unnumbered section). This ensures the proper
%% identification of the section in the article metadata, and the
%% consistent spelling of the heading.

%%
%% The next two lines define the bibliography style to be used, and
%% the bibliography file.
\input{main.bbl}

\bibliographystyle{ACM-Reference-Format}
\bibliography{main}

\end{document}

%% file: main.bbl
%%% -*-BibTeX-*-
%%% Do NOT edit. File created by BibTeX with style
%%% ACM-Reference-Format-Journals [18-Jan-2012].